\documentclass[letter,12pt]{article}
\usepackage[margin=25mm]{geometry}
\usepackage{adjustbox}
\usepackage{amsmath}
\usepackage{amsfonts}
\usepackage{amssymb}
\usepackage{graphicx}
\usepackage{booktabs}
\usepackage{caption}
\usepackage{subcaption}
\usepackage{verbatim}
\immediate\write18{texcount -tex -sum  \jobname.tex > \jobname.wordcount.tex}

\providecommand{\keywords}[1]
{
  \small	
  \textbf{\textit{Keywords---}} #1
}

\title{Robot path planning using deep reinforcement learning}
\author{Miguel Quiñones-Ram\'irez$^{1}$, Jorge R\'ios-Mart\'inez$^{2}$, V\'ictor Uc-Cetina$^{2}$, \\
\small $^{1}$ Universidad Aut\'onoma de Yucat\'an - \small  miguel.aqr99@gmail.com \\
\small $^{2}$ Universidad Aut\'onoma de Yucat\'an - \small \{jorge.rios, uccetina\}@correo.uady.mx \\
}
\date{February 2023}

\begin{document}

\maketitle

\begin{abstract}
Autonomous navigation is challenging for mobile robots, especially in an unknown environment. Commonly, the robot requires multiple sensors to map the environment, locate itself, and make a plan to reach the target. However, reinforcement learning methods offer an alternative to map-free navigation tasks by learning the optimal actions to take. In this article, deep reinforcement learning agents are implemented using variants of the deep Q networks method, the D3QN and rainbow algorithms, for both the obstacle avoidance and the goal-oriented navigation task. The agents are trained and evaluated in a simulated environment. Furthermore, an analysis of the changes in the behaviour and performance of the agents caused by modifications in the reward function is conducted.
\end{abstract}


\section{Introduction}
Navigation competence is essential for mobile robots. To navigate autonomously, a robot must use its sensors' data to identify an optimal or suboptimal path to a target point while avoiding collisions. Generally, a map of the environment is constructed, and then a path planner algorithm is used to find a clear path. However, the task becomes daunting when dealing with sensor noise, tracking errors, and unpredictable surroundings. It also becomes challenging and time-consuming to update the obstacle map accurately, replan the navigation path and predict all possible situations the robot may encounter.

Alternatively, new methods that do not require maps to navigate have been proposed, such as the use of deep reinforcement learning (DRL), introduced by Mnih et al. in 2013 \cite{atarigames}, which has shown the ability to solve complex tasks that require a lot of data processing by combining the reinforcement learning (RL) framework with the artificial neural networks from deep learning (DL). These methods have the advantages of being mapless, having a strong learning ability, lower sensor accuracy dependence, and requiring less human supervision and environment-dependent engineering. 
Contrary to other mapless navigation approaches, which require explicit programming of the robot's behaviour, DRL methods allow the robot to learn the optimal actions to take at each time step by associating them with observations of the environment and a reward signal. Furthermore, unlike pure deep learning methods, they do not require a dataset of labelled samples, which is severely lacking in robotics. Instead, the robot is trained by directly interacting with its environment in a trial-and-error manner. Even when training in the real world proves costly, DRL allows a robot to learn in a simulated environment safely and then transfer the knowledge to a real robot, which is possible because of the generalisation ability of DL models. DRL robotic applications often treat sensor data as a representation of the environment's state, the most commonly used being ranging data, monocular camera images, and depth camera data. Among the sensors used to collect the data, RGB-D cameras are one of the most cost-efficient, lightweight, and information-rich, which allows them to be used for a wide range of applications. As a state representation, RGB images are sensitive to lighting and colour changes, which may be irrelevant to the navigation task. Still, depth images provide geometrical information about the surroundings and are represented as grayscale images, which have been proven to achieve good results in DRL methods applied to different domains. The introduction of deep reinforcement Learning in 2013 by DeepMind \cite{atarigames} demonstrated its potential by training agents that achieved better performance than human experts on Atari games. Since then, notable achievements of DRL methods have been primarily on gaming applications, such as AlphaGo \cite{alphago} winning against the Go champion, AlphaZero \cite{alphazero} beating the champion chess program, and OpenAI Five \cite{openaifive} defeating professional teams in the online game DOTA 2. However, RL approaches to solve real-world problems have been proposed in several domains, including healthcare \cite{healthcare}, analytics \cite{analytics}, language processing \cite{language}, networking \cite{networking}, finances \cite{finances} and robotics \cite{rlrobot}.

Deep reinforcement learning approaches in navigation aim to benefit from learnt skills to solve conventional navigation problems, such as lack of generalisation, the need for fine-tuning or the inability to react in real-time, for applications where mobile robots operate in complex environments. Some of these scenarios include outdoor environments with uneven terrain and noisier sensor readings, dynamic environments where fast reaction times are required, and human environments where collaboration and safety measures are necessary. Deep reinforcement-based applications for navigation have been developed for social robotics, service robotics, unmanned ground vehicles and self-driving cars, among others.

For the autonomous navigation problem, DRL applications are focused on four scenarios, as studied by Zhu et al. in \cite{drlreview}, which include local obstacle avoidance, indoor navigation, multi-robot navigation and social navigation. The applications are usually limited to one of those specific capabilities and are developed by conducting specialised research and adding expert knowledge to favour the convergence of the DRL methods. For that reason, little research has been done on moving from a simpler to a more complicated task. Moreover, few studies analyse the impact of the reward function on the agent's behaviour, as its design is tailored to solve the specific task, and no further comparison is made. Furthermore, the review of Zhu et al. \cite{drlreview}, the survey of DRL algorithms for autonomous vehicles of Ye et al. \cite{drlalgorithm}, as well as the related works reviewed, indicate that among the most commonly used DRL algorithms are the Deep Q Networks (DQN) \cite{atarigames}, Double DQN (DDQN) \cite{doubledqn}, Dueling DDQN (D3QN), Asynchronous Advantage Actor-Critic \cite{a3c}, Proximal Policy Optimization \cite{ppo} and Deep Deterministic Policy Gradients \cite{ddpg}. However, since their introduction, improvements have been proposed in each algorithm's family of RL methods that lead to new state-of-the-art performances in their benchmark domain, such as continuous control or Atari Games. This means that more modern DRL methods could also improve the results in autonomous navigation-related tasks. In the present research, those problems are studied by training and evaluating different DRL agents in obstacle avoidance and goal-oriented navigation tasks, which were designed considering the challenges presented in the previous reviews. As mentioned by Zhu et al. in \cite{drlreview}, the term $mapless$ used to describe DRL-based navigation systems in this work refers to the use of lightweight localisation solutions, such as GPS and WiFi, to obtain the relative position of the goal point without a global map. Although the training environments were designed based on the conditions of an indoor navigation scenario, the goal-oriented navigation task is referred to as such due to its focus on reaching a goal rather than the complexity of the environment.

\begin{figure}
    \centering
    \includegraphics[width=8cm]{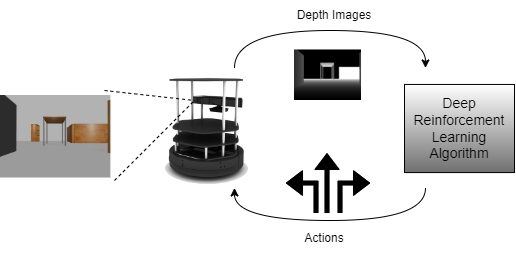}
    \caption{An intuition of the RL framework applied to a robot. Based on an observation of the environment, the robot is given the optimal action to take.}
    \label{fig:idea}
\end{figure}

This article introduces a mapless deep reinforcement learning approach to solve the autonomous navigation problem in indoor and static simulated environments using depth images. It focuses on analysing the different data required to train agents for obstacle avoidance and goal-oriented navigation tasks, studying the effect on their behaviour and performance by modifying the reward signal and changing the algorithm used. The proposed approach is implemented in the open-source mobile robot Turtlebot2 \footnote{https://www.turtlebot.com/turtlebot2/}, by using the Robotic Operating System (ROS) \footnote{http://wiki.ros.org/} as the robotics framework and Gazebo \footnote{https://gazebosim.org/home} as the robotics and physics simulator. However, the DRL framework can be applied to different mobile robots and using other robotic simulators as long as it is possible for the robot to perform the designated actions and the necessary sensory data is available. An initial idea about how a robot follows an RL approach in a navigation task is shown in Fig. \ref{fig:idea}.

\section{Autonomous Navigation}

Autonomous navigation is one of the biggest challenges for a mobile robot. A robot must succeed at four building blocks to navigate autonomously: perception, localisation, cognition, and motion control \cite{robotbook}. Perception requires taking measurements, using different sensors, and extracting meaningful information from those measurements. Localisation involves determining the robot's absolute position in space and relative position concerning its goal and the obstacles. Cognition includes decision-making and its execution to achieve the highest-order goals. Moreover, motion control modulates the robot's motor outputs to achieve the desired trajectory. For a mobile robot, the navigation competence is required for its cognition. Given partial knowledge about its environment and a goal position, navigation encompasses the capability of the robot to act based on its knowledge and sensor values to reach the goal as efficiently as possible \cite{robotbook}. However, obstacle avoidance and path planning competencies are also required for autonomous navigation. There may need to be more than a behaviour or reactive navigation \cite{robotbook2} for a mobile robot to reach a distant goal. Likewise, a plan might only be accomplished if the robot can react to unforeseen events. For that reason, modern navigation methods combine both competencies, sensor data and a map, to create a plan, execute it and make adjustments during motion.

\subsection{Obstacle Avoidance}

Obstacle avoidance requires controlling the robot's trajectory to prevent collisions by making decisions based on sensor readings \cite{robotbook}. Unlike path planning, it is reactive and considers only a few steps ahead when making decisions. One of the simplest obstacle avoidance algorithms is the Bug Algorithm, which follows the contour of each obstacle to circumnavigate it. The robot stops its movement around the obstacle when it finds a minimum distance point towards its destination or a slope equal to its original one, meaning that it requires at least the robot's localisation. \cite{robotbook2}
As an obstacle avoidance approach with access to knowledge of its environment, the Bubble Band technique generates a subset of the free space around a robot that can be travelled without collision using a map and range information. A string of these so-called bubbles is later used to indicate the trajectory to the goal position \cite{robotbook}. For more robustness, the Vector Field Histogram (VFH) technique generates a 2D polar histogram of the environment around the robot based on its sensor readings. Then it converts it into a 1D polar histogram, where the x-axis represents the angle at which an obstacle was found and the y-axis the probability of it being there. Then, a path is chosen based on the obstacle density, the robot's alignment with the goal, and its steering angle \cite{obstacles}. Dynamic Window Approach (DWA) is a method that goes a step further by considering the robot's kinematics constraints to select an appropriate combination of linear and angular velocities that allows it to avoid running into obstacles. Given the current robot's speed, the local version of DWA selects a set of tuples of linear and angular velocities which can be reached within the next sample period, also known as the dynamic window. Then, the set is reduced to only those which allow it to stop before hitting an obstacle, given by an objective function, and selects the best tuple based on an objective function. The global version of DWA considers the distance to a goal in the objective function, allowing it to have a more long-termed view \cite{robotbook}. 
Fuzzy Logic Controllers are an alternative approach that uses ambiguous and noisy data to make decisions by selecting a proper action based on a set of rules that model a reasoning capability. They improve the performance of mobile robots in complex environments, but at the cost of the complexity that entails designing the set of heuristics \cite{obstacles2}. For a more detailed explanation of obstacle avoidance methods, the work of Shitsukane et al.\cite{obstacleavoidancesurvey} can be consulted.

\subsection{Path Planning}

Path planning is defined as the problem of finding a sequence of valid configurations to move from a starting position to a goal position and requires a model of the environment transformed into a discrete map. However, most mobile robots use differential-drive systems, which impose nonholonomic constraints on their configuration. Furthermore, when they are on the ground, their path planning is often considered to take place in a 2D representation of the environment \cite{robotbook}. For that reason, typical representations of the environment include grid maps, metric maps, and topological maps. Path planning is classified by the environment and the knowledge that the robot has about it. If the robot has complete knowledge about its environment, it is known as a global path planning problem, in which the planner has to compute an optimal path to the goal. In contrast, a local path planner uses sensor readings to constantly obtain information about the robot's surroundings and follow a path while avoiding obstacles. Local path planning is associated with obstacle avoidance, while global path planning includes graph-based and potential field-based methods \cite{pathplanning}.

Graph Search methods rely on using a map that indicates the free and occupied space in the environment to build a graph and compute a solution. Then, graph search algorithms can be used to find a path, such as breadth-first search, depth-first search, Dijkstra's algorithm, or the A* algorithm. Among these, the A-star algorithm stands out for its consistency, speed, and ability to find the optimal solution at the cost of being computationally more expensive and requiring a heuristic function and path cost function, which may be difficult to define in some cases. Rapidly Exploring Random Trees (RRT) is also a fast alternative that does not require a heuristic function, and its lack of solution optimality was addressed by RTT*. Potential Field path planning methods define forces that either attract the mobile robot towards the goal or repel it from certain positions, such as obstacles. The environment is modelled based on the forces, and the robot is a point under its influence. As long as the robot can localise its position concerning the potential field, it can compute its following action based on the forces surrounding it. A more in-depth analysis of the path planning problem can be seen in the work of Sánchez-Ibáñez et al. \cite{pathplanningsurvey}.

\subsection{Robot Navigation Systems}

Autonomous navigation systems require a path-planning method, an obstacle-avoidance approach, and a localisation method to provide the necessary information for both. Sensor data may be used directly in some cases. However, without knowledge about its position relative to the goal, a mobile robot is limited to reactive behaviour, following a predetermined path, or chasing short-termed goals based on its sensor range \cite{visualnavigation,robotbook2}. A broad classification of autonomous navigation techniques is whether they are used indoors or outdoors, as well as regarding their consideration of dynamic obstacles. Indoor environments have their working space clearly defined and the surface area physically delimited, and the boundaries are easily identifiable by the robot's path planning and obstacle avoidance algorithms. The limited space and predominance of flat surfaces favour map-based and map-building systems because the robustness and reliability outweigh the computing cost when the resources are available. On the other hand, outdoor navigation systems must deal with uneven terrain, noisier sensor readings due to environmental causes, and more uncertainty about the robot's whereabouts due to its unstructured environments. Navigation in dynamic environments is more complex, requiring not only estimating the position of static obstacles and boundaries but also constantly being on the lookout for movement or any other indication that an obstacle may be headed toward the robot's path. Dynamic navigation systems have a broader selection of applications but require fast updates. However, the inclusion of dynamic obstacles is beyond the scope of this work.

Indoor navigation techniques can also be categorised as map-based, map-building-based, or mapless, depending on the source of the goal-related information they use. Map-based approaches must be provided with a representation of the environment built by a different system beforehand. Map-building techniques can compute the model of the environment themselves and use it subsequently as a source of information. Mapless methods rely on their sensors alone, primarily on visual data, to infer knowledge about their goals' position based on the features detected during motion \cite{visualnavigation}.
RL enables a nature-inspired approach, in which robots learn the optimal behaviour to fulfil a task by interacting with their environment instead of being programmed explicitly. Combined with the advancements in the DL field, it allows them to extract meaningful features from their environment and decide which actions to take without an explicit rule. A DRL-based approach allows a robot to behave similarly to other mapless methods and train specific tasks that complement or improve existing navigation systems.

\subsection{Conventional Navigation}

In most navigation problems, the robot does not have access to an accurate map of the environment, and the most popular approach to solve them is by using a map-building system. For that reason, navigation is also referred to as the combination of localisation, map-building, and path planning. In that case, the standard technique is to perform the three tasks simultaneously, known as Simultaneous Localization and Mapping (SLAM) \cite{visualnavigation}. Different algorithms have been proposed to solve the SLAM problem, with the most commonly used being laser, sonar, or visual sensors. The SLAM problem has been studied for many years and has become the industry standard technique to solve navigation problems due to its robustness and reliability, despite the cost of computing and updating a map \cite{slam}. 

Even the industry and academic most popular robotics framework, the Robot Operating System (ROS) \footnote{http://wiki.ros.org/}, describes a default navigation system like a map-building system, which requires the computation of a map through odometry and sensor data, and the use of global and local path planners \cite{rosbook}.  Commonly used algorithms in the navigation stack \footnote{http://wiki.ros.org/navigation} include GMapping, Adaptive Monte Carlo Localization, $A*$ for the global path planner, and DWA for the local path planner and obstacle avoidance. RL offers a mapless approach for solving navigation tasks, a better generalisation capability combined with Deep Learning, and the ability to perform complex behaviours without engineering them. The RL framework allows for versatility and is not limited to using distances as observations and velocities as outputs but can be trained with different data depending on the task. Furthermore, its learning capability is not limited to pre-established rules. It learns to associate a given observation of the environment with the optimal action to fulfil the task as efficiently as possible. Also, the learning process is performed before the robot is put in motion, allowing simulators to be used as a safe training space. It also eases the load on the robot during movement because it already knows what action to take in each scenario.

\section{Reinforcement Learning}

Reinforcement Learning (RL) is one of the three essential Machine Learning (ML) paradigms. RL aims to enable an agent to learn the optimal behaviour to accomplish a task by repeatedly interacting with its environment \cite{rlbook}, differing from supervised learning and unsupervised learning, which rely on given data sets.

The main elements of an RL problem are the agent, its possible actions, the environment it belongs to, the state of the environment at any given time, the reward the agent receives from the environment, and a policy that defines the agent's behaviour. The agent is associated with the model that carries out the decision-making progress and learns; it does not refer to a physical entity. The actions are the set of decisions that the agent can take to interact with its environment. The environment generally refers to anything that the agent cannot arbitrarily change. At the same time, a state is the complete description of the environment at a given time. The reward signal is a numerical value that indicates how well the agent performed and is perceived by the environment on each time step. Finally, the policy is a rule used by the agent for its decision-making process, which maps the states perceived from the environment to actions to be taken when being in them. 

\subsection{Markov Decision Processes}

Markov decision processes (MDPs) are used to formally define the interactions between a learning agent and its environment and as the mathematical foundation of an RL problem. An MDP is a system described by the set of states $ S $, the set of actions $ A $, the reward fucntion $ R : S \times A \times S \rightarrow R $ and a transition probability function $ P : S \times R \times S \times A \rightarrow [0,1] $\cite{rlbook}; and also obeys the Markov property
$$ p(s',r|s,a) = Pr\{S_t=s',R_t=r|S_{t-1}=s,A_{t-1}=a\} $$
which establishes that future states only depend on the most recent state and action. MDPs are a formalization of sequential decision-making, where actions influence future states and rewards, and by using them, it is possible to predict all future rewards and states. When the agent has access to the transition probability function, also referred to as the model of the environment, it is possible to use model-based RL methods, which rely on state transitions and reward predictions to plan. However, in most cases, a ground-truth model of the environment is not available, and the agent must follow a model-free approach to learn purely from experience by associating states to actions through some computation. 

\subsection{Returns and Episodes}

\begin{figure}
    \centering
    \includegraphics[width=8cm]{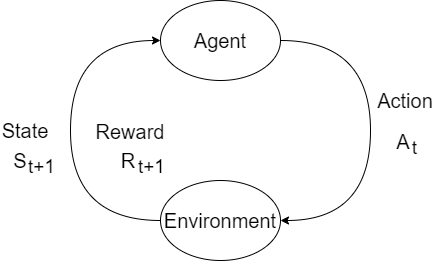}
    \caption{The agent-environment interaction. The agent observes the current state, selects an action, receives a reward and an observation of the new state.}
    \label{fig:rlinteraction}
\end{figure}

At each time step $ t $, the agent observes the current state $ S_t = s \in S $ of the environment, proceeds to take an action $ A_t = a \in A $, and is provided with a reward $ R_{t+i} $ by the environment. Then, the environment transitions to a new state $ S_{t+1} = s' $ and the cycle is repeated, as shown in Fig. \ref{fig:rlinteraction}. By looking for correlations between states, actions and rewards, the agent learns to perform its task efficiently \cite{rlbook}. 

The agent's goal is to maximize the cumulative reward it receives, also known as the return, which can be defined as the sum of the rewards at each time step:

$$ G_t = R_{t+1}+R_{t+2}+R_{t+3}+...$$

To prevent an infinite amount of return, the concept of discounting is introduced, and the discounted return is defined as:

\[G_t = R_{t+1}+\gamma R_{t+2}+\gamma^2R_{t+3}+... = \sum^\infty_{k=0} {\gamma^{k}R_{t+k+1}}\]

where $ \gamma $ is the discount rate and determines the value of the future rewards, $ 0 \leq \gamma \leq 1 $. 

However, in many cases, the agent-environment interaction can be broken down into sub-sequences, called episodes, with a final time step $ T $. Each episode ends in a terminal state followed by a reset to a starting state.

\subsection{Policies and Value Functions}

A policy maps states to probabilities of selecting each possible action. When an agent follows a policy $ \pi $, then $\pi(a|s)$ is the probability of performing the action $ a $ when at the state $ s $.
The goal of an RL algorithm is to discover an optimal policy $\pi^*$ that prioritizes the best action to take at each state, so as to maximize $ G $ \cite{rlbook}.  For that reason, it is useful to know how valuable a state is. 

A value function $ v_\pi(s) $ is defined the expected return when starting in a state $ s $ and subsequently following a particular policy $\pi$:

\[v_{\pi} = E_{\pi}[G_t|S_t=s] \]

Similarly, an action-value function $ q_{\pi} $ is defined as the expected return when starting from $ s $, taking the action $ a $, and thereafter following the policy $\pi$:

\[q_{\pi} = E_{\pi}[G_t|S_t=s,A_t=a] \]

A policy $ \pi $ con be compared to a different policy $ \pi' $ given their expected returns

\[ \pi \ge \pi' \textrm{ if and only if } v_{\pi}(s) \ge v_{\pi'}(s) \textrm{ for all } s \in S \]

The policy that is better than or equal to all others is considered the optimal policy $ \pi^* $ and is associated with an optimal state-value function $v_*$ or an optimal action-value function $q_*$, defined as
$$v_*(s)=\max_{\pi} v_\pi (s)$$
$$q_*(s,a)=\max_{\pi} q_\pi (s,a)$$

Both types of value functions follow a consistency condition, the Bellman Equation, which expresses the relationship between the value of a state and the value of its possible successor state. The Bellman optimality equation for $v_*$ and  $q_*$ are

\[v_{*}(s) = max_a E[R_{t+1}+\gamma v_*(S_{t+1})|S_t=s,A_t=a] \]
\[q_{*}(s,a) = E[R_{t+1}+\gamma \max_{a'} q_*(S_{t+1},a')|S_t=s,A_t=a] \]

Depending on the RL method, there are different approaches to reaching optimal behaviour. Policy-based or policy optimization methods directly approximate the optimal policy of the agent, while value-based methods learn to estimate it through the use of value functions or state-action functions.

Also, off-policy RL methods use a behaviour policy to select an action and explore the environment different from the target policy that is learnt and improved. Contrary to on-policy methods, where the target and behaviour policy are the same. 
Online methods that update their parameters while observing a stream of experiences, can use two different policies and update them separately. While offline methods commonly optimise only the target policy, and copy its parameters into the behaviour policy, as a less memory-consuming approach, by storing and using experiences at different points of time during the training, through the use of large buffers.

\subsection{Temporal-Difference Learning}

Temporal-Difference Learning refers to a class of model-free, value-based methods, which update their estimate of the value function based on previous estimates without waiting for a final outcome, also known as bootstrapping.
Given some experience following a policy $\pi$, TD methods update their estimate $V$ of $v_\pi$ at each time step $t+1$ by using the observed reward $R_{t+1}$ and the estimate $V(S_{t+1})$ \cite{rlbook}

$$V(S_t) \leftarrow V(S_t)+\alpha [R_{t+1}+ \gamma V(S_{t+1})-V(S_t)]$$

The most basic type of TD method is called the one-step TD because the target for the TD update is calculated using the value and reward of only the next time step. The quantity in brackets in the one-step TD is also called the TD error because it measures the difference between the estimated value of $S_t$ and the better estimate $R_{t+1}+\gamma V(S_{t+1})$, available one step later. As long as the step-size parameter $\alpha$ is sufficiently small, one-step TD converges deterministically to a single answer.

The advantages of TD methods over others are that they do not require a model of the environment and do not need to wait until the end of the episode to learn.

\subsection{Q-Learning}

Q-learning is an off-policy TD method and one of the most popular Reinforcement Learning algorithms. It is defined by the update to the action-value function:

\[Q(S_t,A_t)\leftarrow Q(S_t,A_t) + \alpha[R_{t+1}+ \gamma \max_{a}Q(S_{t+1},a)-Q(S_t,A_t)]\]

And to approximate the optimal action-value function $q_*$, the agent must visit, store in a tabular manner, and update all the state-action pairs, also known as Q-values, for the action-value function Q \cite{qlearning}.

\section{Deep Reinforcement Learning}

\begin{figure}
    \centering
    \includegraphics[width=8cm]{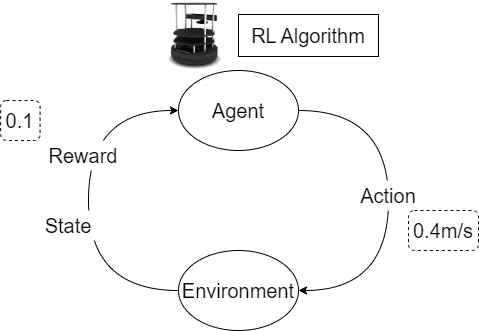}
    \caption{An example of a robot described in the RL framework. The state is still to be defined.}
    \label{fig:barebones}
\end{figure}

The previously described framework may be used to apply an RL approach to a robotics problem. In the case of autonomous navigation, the robot can be seen as the agent, its linear and angular velocities as the actions and the reward should incentive the robot to evade obstacles or move closer to its goal, as shown in Fig. \ref{fig:barebones}. However, the challenge lies in defining an appropriate state that provides enough information for the robot to fulfil its task, especially for robots that operate in a three-dimensional space.

In 2013, Kober et al. published a survey \cite{rlrobot} about the challenges and successes of Reinforcement Learning in Robotics, and one of the main challenges is the "Curse of Dimensionality". This holds, especially for robotics, where multiple sensor readings, degrees of freedom or images are needed to describe the robot's state space. However, in the same year, Google DeepMind proposed a novel algorithm, Deep Q Networks (DQN) \cite{atarigames}, by combining the traditional Q-learning method with a Neural Network, which vastly outperformed all previous methods at playing Atari games with RGB images as inputs. This work started the trend of combining RL methods with Neural Networks from the DL field, which became a subfield known as Deep Reinforcement Learning.

When designing an agent that uses depth images as states, the improved computational capabilities and robustness of the DRL are needed for the agent to be able to process the data and extract meaningful features that allow it to differentiate and evaluate each state.

\subsection{Neural Networks}

Artificial Neural Networks, or simply Neural Networks (NNs), are computing models based on a collection of connected nodes known as neurons, used to approximate high-dimensional and non-linear functions. The neurons are aggregated into layers, where different transformations are performed and associated with the weights adjusted for the network to learn. The neurons are inspired by the brain cells of the same name, and their design is based on the perceptron, introduced by Frank Rosenblatt in 1958 \cite{perceptron}. Each neuron's inputs are weighted, summed and added a bias before being passed through an activation function that applies a non-linear transformation, which is the main reason why they perform well in different applications.

Each NN has an input layer, where data is introduced, an output layer, where a prediction is given, and many hidden layers in between, where the values are computed. The more hidden layers are used, the better the capability of the network to abstract meaningful information from the data to make better predictions. For that reason, the term $deep$ originates from using a larger amount of hidden layers, which was possible due to the increase in available computing power and memory, contrary to the earlier $shallow$ networks.

The most basic type of neural network is a feedforward neural network \cite{dlbook}, or multilayer perceptron, where each layer is composed of many neurons, and their output is connected to the input of the next layer. The layers of these types of NNs are known as feedforward, fully connected or linear layers due to their sequential nature and because all of the neurons are connected to the next layer. The number of neurons and the activation function for each layer can be modified, with the most commonly used being the $reLu$, $tanh$, $sigmoid$ and $softmax$ functions.

A specialised type of NN for processing data that has a grid-like shape is known as the Convolutional Neural Network (CNN) \cite{dlbook}, and its most popular use is for processing images. CNNs have layers that perform a convolution instead of a matrix multiplication, known as convolutional layers. The convolution requires sliding a  kernel, a smaller array of parameters, along the input matrix of the slayer and performing a dot product in small windows of features, reducing the output data size. The size of the kernel, the number of kernels, the amount of stride that the kernel slides, and whether the input features are padded to keep their size after the operation, among other features, can be tuned for each convolutional layer. The convolution operation allows extracting high-level features from images, such as edges and colour. It performs better predictions, and the popularity of this type of network increased thanks to the results of trained models such as AlexNet, presented in \cite{alexnet}, and ResNet, proposed in \cite{resnet}.

\subsection{Deep Q-Networks}
 
The Q-Learning algorithm's limitations to store and approximate the value of all state-pairs when the number of combinations is increased was addressed by Mnih et al. in \cite{atarigames}. They proposed an approach called Deep Q Networks that combined the Q-learning algorithm with Neural Networks.

The core idea was to approximate the Q-values using a Deep Neural Network (DNN) instead of storing them in a tabular manner. To that end, the value function was parametrised as $Q(s,a;\theta_i)$ by using the neural network's weights $\theta$ at each time step $i$ \cite{naturedqn}. The Q-learning update becomes the loss function to train the neural network. The loss is given by:

\[L(\theta) = E_{s,a,r,s'}-U(D)[(r+ \gamma \max_{a'}Q(s',a';\theta_i) - Q(s,a;\theta_i))^2]\]

Where, at each time step $t$, the agent's experiences $e_t=(s_t,a_t,r_t,s_{t+1}$ are stored in an experience replay $D_t={e_1,...,e_t}$, and a mini-batch of experiences $(s,a,r,s')$ is drawn uniformly at random, $U(D)$ , to perform the update.

This method outperformed most state-of-the-art methods at Atari games without prior knowledge and by using raw images and established the beginning of Deep Reinforcement Learning.

\subsection{Double DQN}

One disadvantage of the Q-learning algorithm, as evidenced by van Hasselt, is the overestimation of action values due to a positive bias from using the maximum action value as an approximation for the maximum expected action value. A double estimator method was proposed to decouple the action selection process from the evaluation and eliminate the bias, resulting in an underestimation of action values. Furthermore, van Hasselt et al. \cite{doubledqn} extended the idea for its use in parametric equations and the DQN algorithm, proposing the variant Deep reinforcement learning with Double Q-learning (Double DQN or DDQN) by using two Neural Networks with different sets of weights. The main neural network picks the best next action $ a' $ among all the available, and then the target neural network evaluates the action to know its Q-value. While the main neural network's weights are updated normally, the target neural network is updated every so often with a copy of the main neural network's weights. The Bellman equation in this algorithm has the shape:

\[Q(s,a;\theta)=r+ \gamma Q(s',argmax_{a'}Q(s',a';\theta); \theta ')\]

\subsection{Prioritized Experience Replay}

The Experience Replay, introduced by Lin \cite{experiencereplay}, helped online RL methods to break the temporal correlations of the updates and to prevent the loss of rare experiences by mixing more and less recent experiences and allowing them to be used multiple times. However, experiences are sampled uniformly at random, without regard for each experience's value. The Prioritized Experience Replay (PER), proposed by Tom Schaul et al. \cite{per}, focuses on the effective use of the replay memory for learning by prioritising transitions which may be more valuable for the agent but rarely occur. The TD error $\delta$ is used as a criterion to measure the importance of each transition by indicating how unexpected each transition is because it compares how far the value is from the next bootstrap estimate. However, purely choosing the experiences with the most TD error would lead to over-fitting. Therefore, a stochastic sampling method was proposed that interpolates greedy prioritisation and uniform random sampling.

So each transition $i$ is given a priority value
$$p_i = |\delta_i|+\epsilon$$
where $\epsilon$ is a small positive constant that prevents a transition from not being visited, such that $p_i>0$. And the probability of sampling each transition $i$ is given by
$$P(i) = \frac{p_i^\alpha}{\sum_k p_k^\alpha} $$
where the $\alpha$ determines how much prioritization is used, with $\alpha=0$ corresponding with the uniform case. And to prevent the bias toward high-priority samples introduced by the change of distribution in the stochastic updates, importance-sampling (IS) weights are used
$$w_i=\left(\frac{1}{N}\frac{1}{P(i)}\right)^\beta$$
where $N$ is the size of the replay buffer, and the Q-learning update is performed using $w_i\delta_i$ instead of $\delta_i$.
The hyperparameter $\beta$ controls how much the IS weights affect learning and is linearly annealed from an initial value $0 < \omega < 1$ to 1.

\subsection{Dueling Network}

The Dueling Network architecture, proposed by Xie et al. \cite{duelingdqn}, splits the Q-values between the value function V(s) and the advantage function A(s, a). The first one estimates the reward collected from the state $'s'$, while the second one estimates how much better one action is compared to the others.

The Q-value is defined by:

\[Q(s,a)=V(s)+A(s,a)\]

For that reason, the Dueling Network has two streams to separately estimate state values and the advantages for each action and combine them to output Q-values for each action. To prevent the Q-value equation from being unidentifiable, the advantage function estimator is forced to have zero advantage at a chosen action:

\[Q(s,a)=V(s)+ (A(s,a)- \frac{1}{|A|} \sum_{a'} A(s,a))\]

Because the dueling architecture shares the same input-output interface, it can be combined with other Q network-based architectures. One of the algorithms which significantly improved when combined with a dueling architecture is the DDQN, and such combination is often referred to as Dueling Double DQN or D3QN.

\subsection{Multi-step Learning}

The idea of multi-step learning, or originally known as n-step Bootstrapping \cite{rlbook}, comes from the comparison between TD methods and other RL methods, such as the Monte Carlo (MC) methods. Whereas most TD methods bootstrap their estimations over every time step, MC methods do so only at the end of each training episode. Therefore, a middle ground was proposed in which it is possible to bootstrap over a length of time in which significant state changes have occurred, effectively leading to faster learning.
The truncated n-step return from a given state $S_t$ is defined as 

\[R_t^n= \sum^{n-1}_{k=0} {\gamma^{k}_t R_{t+k+1}}\]

And the multi-step variant of the DQN loss is defined as
$$(R_t^n+\gamma_t^n \max_{a'} q_\theta (S_{t+n},a')-q_\theta (S_t,A_t))^2$$

\subsection{Distributional Reinforcement Learning}

Bellemare et al. \cite{distributionalrl} proposed a method to model the full distribution of returns instead of only the expectation, which leads to better approximations and more stable learning. The returns' distribution is modelled using a discrete distribution parametrised by $ N \in \mathbb{N^{+}} $ and $ V_{MIN}$,$ V_{MAX} \in \mathbb{R} $, with probability masses placed on a discrete support $z$, where $z$ is a vector of $N$ atoms, considered as the canonical returns, defined by

$$z^i = V_{MIN} + i (\frac{V_{MAX}-V_{MIN}}{N-1})$$

for $i \in {1,...,N}$. With the probability mass of each atom

\[p_\theta^i(s,a) = \frac{e^{\theta_i(s,a)}}{\sum e^{\theta_j(s,a)}}\]

such that the approximating discrete distribution $d$ at time $t$ is given by

$$d_t = (z,p_\theta(s,a))$$

A variant of Bellman's equation is used to learn the probability masses.
The Bellman operator $T^\pi$ is defined to describe the contraction by $\gamma$ and shift by the reward of the future estimation, to get the current value during the policy evaluation. The Bellman Equation

\[Q^\pi(s,a) = \mathbb{E}R(s,a)+\gamma \mathbb{E}_{P,\pi}Q^\pi(s',a')\]

can be rewritten using the Bellman operator

\[T^\pi Q(s,a) = \mathbb{E} R(s,a)+\gamma \mathbb{E}_{P,\pi} Q(s',a')\]

The Bellman operator $T^\pi$ is further proved to converge to a unique return distribution by using a metric between cumulative distribution functions, known as the Wasserstein Metric. Denoting the return as $Z$ and the return distribution as $Z^\pi$, the convergence of $Z$ is studied by applying the Bellman operator, as

\[T^\pi Z(s,a) = R(s,a)+\gamma P^\pi Z(s',a').\]

However, when extending the idea to the Bellman optimality operator $T$

\[T Q(s,a) = \mathbb{E}R(s,a)+\gamma \mathbb{E}_{P}max_{a' \in A}Q(s',a'),\]

it can only be proved that $T$ converges to a set of optimal return distributions.

Furthermore, applying $T$ to $Z$ cannot be computationally done without applying the $argmax$ function to the expectation of the future value.

\[T^*Z(s,a) = R(s,a)+\gamma Z(s',max_{a' \in A}E[Z(s',a')])\]

When applying the Bellman update $TZ_\theta$ to the parametrisation $Z_\theta$, the supports are almost always disjointed. To fix this, and considering an issue with the Wasserstein loss when sampling from transitions, the sample Bellman update $\tilde{T}Z_\theta$ is projected onto the support of $Z_\theta$, reducing the update to a multi-class classification.

\subsection{Noisy Networks}

One of the key challenges of RL methods is maintaining a balance between exploration and exploitation. Traditional exploration heuristics rely on random perturbations of the agent's policy, such as $\epsilon$-greedy, probabilities, or intrinsic motivation terms added to the reward signal, to encourage new behaviours. However, these methods are not easily applied with neural networks or rely on a metric chosen by the experimenter.
For that reason, Fortunato et al. \cite{noisynet} proposed NoisyNet, an approach where perturbations of a neural network's weights are used to drive exploration.
The number of parameters in the linear layer of the neural network is doubled and allows for different learning rates at the state space.
For a linear layer of a neural network
$$y=wx+b$$
the corresponding noisy linear layer is defined as
$$y=(\mu^w + \sigma^w \odot \epsilon^w)x + \mu^b + \sigma^b \odot \epsilon^b$$

where the parameters $\mu^w, \sigma^w, \mu^b, \sigma^b $ are learnable and 
$\epsilon^w, \epsilon^b$ are noise random variables originating from either an Independent Gaussian noise or a Factorised Gaussian noise.

\subsection{Rainbow}

\begin{figure*}
    \centering
    \includegraphics[width=1\textwidth]{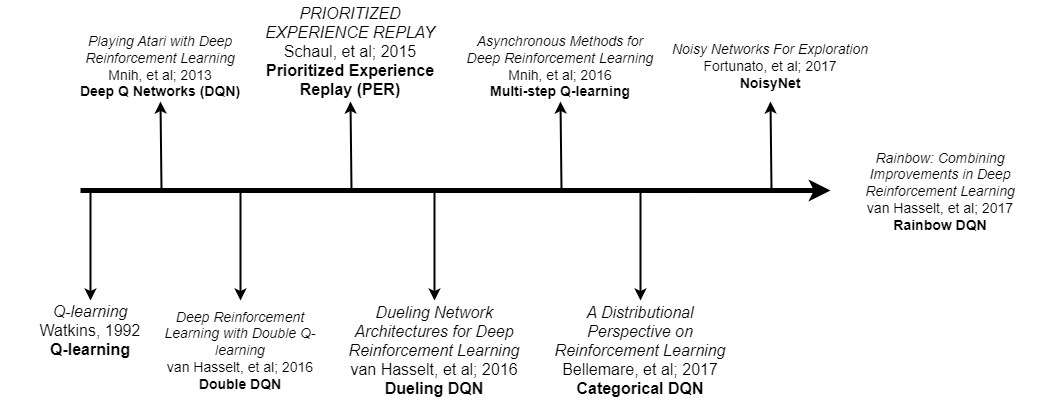}
    \caption{Rainbow DQN components. The combination of independent improvements resulted in a better performance than the baseline DQN.}
    \label{fig:rainbow}
\end{figure*}

All previous improvements to the original DQN algorithm were made independently, as illustrated in Fig. \ref{fig:rainbow}. Hessel et al. \cite{rainbowdqn} proposed that each extension addressed a distinct concern and that they could be combined to improve the performance of the DQN algorithm. The distributional loss is replaced with a multi-step variant. A shift by the truncated n-step discounted return is considered at the value $S_{t+n}$ in the Bellman operator instead of the original shift by the return. The target distribution is defined as 

$$d_t^{(n)}=(R_t^{(n)}+\gamma _t^{(n)}z, p_{\tilde{\theta}}(S_{t+n},a*_{t+n})$$

where the greedy action $a*_{t+n}$ is selected by the online network to bootstrap, and the target network evaluates the use of said action.

The resulting KL loss is:

$$D_{KL}(\phi_z d_t^{(n)}||d_t)$$

which can be used to compute the priority values of the PER as a more robust and efficient alternative to the TD error. The neural networks follow the dueling network architecture but are adapted for use with return distributions. And finally, all the linear layers are replaced with noisy linear layers.

\section{Deep Reinforcement Learning for Navigation}

In mapless navigation systems, there isn't an available representation of the environment; the robot perceives the environment as it navigates and must be able to recognise objects, landmarks or any similar type of information that allows it to infer knowledge about where its goal is located. Most of these systems use visual information, primarily the first-person-view image, and perform some reactive behaviour as they process the incoming data \cite{maplessnav}. 

Optical Flow methods use a sequence of images to estimate the motion of objects and features. Velocities perceived are used for the robot's decision-making, always preferring to move in the direction of less change. This is also the main disadvantage of these methods \cite{opticalflow}. Appearance-based methods store and memorise images of the environment and associate them with certain relative localisation to the goal, allowing the robot to perform the correct motion control. However, labelling the desired images and developing the appropriate criteria may be difficult and time-consuming \cite{opticalflow}.  Feature tracking-based methods rely on detecting features and motion from the elements in the environment and, based on that information, estimate the robot's trajectory and motion \cite{visualnavigation}. Object recognition is more symbolic and can detect features rather than memorising precise objects or positions. Deep learning approaches are very similar in the sense that neural networks are trained with many images to identify features of the objects in the environment \cite{dlnav}.

All the aforementioned methods are limited to one task except the DL-based. All of them require the use of labelled images that indicate the desired motion at a specific place or the landmark it represents, which can be very costly to produce. On the contrary, RL agents can be trained for different tasks and allow simulated environments to safely and efficiently train an agent before transferring it to a real-life robot, reducing the computational load needed to learn the task. Finally, technological advances allow the recreation of more realistic and complex scenarios and accelerate learning.

Choices of DRL algorithms in robotics include different variations of DQN, and policy search methods, such as Proximal Policy Optimization \cite{ppo} (PPO), Asynchronous Advantage Actor-Critic \cite{a3c} (A3C) and Deep Deterministic Policy Gradients \cite{ddpg} (DDPG). In the case of mobile robots, different tasks have been accomplished using DRL methods, the most common being obstacle avoidance and navigation. However, more complex tasks can be performed depending on the information provided to the agent. A summary of the related works can be seen in Table \ref{alternaterl}.

For the obstacle avoidance task, Lei Tai and Ming Liu \cite{indoordqn} implemented a DQN agent trained to explore indoor environments by using depth images as the states and a CNN pre-trained with real-world samples. Linhai Xie et al. \cite{monoculard3qn} combined a CNN trained to predict depth images with a D3QN agent to propose an approach that uses only monocular RGB vision as input. They also showed that the D3QN model outperformed a vanilla DQN model on both training speed and performance. Patrick Wenzel et al. \cite{visionbasedobstacledrl} also used a NN to predict depth images based on RGB images and implemented three different agents to solve obstacle avoidance in circuit-like environments: a PPO agent with a discrete action set, a PPO agent with a continuous action set and a DQN agent. They concluded that the PPO agent with discrete actions outperformed the other two agents and that depth images yielded better results than RGB and grayscale images.

For the goal-oriented navigation task, Xiagang Ruan et al.\cite{navd3qn} implemented a D3QN agent that successfully navigates autonomously by using depth images and the distance to the goal as a state. Changan Chen et al. \cite{crowdsarl} presented an LSTM network that models Human-Robot and Human-Human interactions, using the DRL framework, for navigation towards a goal in a crowded environment. Yuke Zhu et al. \cite{thornav} trained an A3C agent in a self-developed physics engine, which could generalise across targets and scenes. Two RGB images were used for the state representation, one from the agent's perspective and another that shows the target, and were embedded by a CNN before being passed to the agent. Liulong Ma et al. \cite{usingrgb} compared two DRL agents, DQN for a discrete action space and PPO for a continuous action space, to perform a mapless navigation task by using a Variational Autoencoder to encode RGB images and appending them with target related information. The PPO model outperformed the DQN model in both performance and training time and also got better results in its environment than the benchmark. Cimus Reinis et al. \cite{goobstacle} proposed a DDPG agent that combined a stack of depth images with the polar coordinates between the robot and the goal as the state and with a reward based on the robot's velocity. They performed successful experiments on simulated environments as well as real-world scenarios.

However, other works involve different navigation-related tasks, such as Pararth Shah et al.\cite{follownet}, which combined a DQN agent with a Recurrent Neural Network to map natural language instructions, and visual and depth inputs to actions. Wenhan Luo et al. \cite{14trackingrl} developed an A3C agent for a mobile robot, combined with a ConvLSTM NN, that takes RGB frames as inputs and produces both camera control and motion control signals as outputs. Their agent could resume tracking after losing the target and was successfully transferred to real-world scenarios. Placed and Castellanos \cite{d3qnslam} developed a D3QN agent capable of performing active SLAM with less intensive computation by using laser measurements and designing the reward function based on a formulation of the active SLAM problem.

While most studies specialise in a task and propose a specific reward function and state representation to fulfil it, the work presented analyses the challenge involved in going from a simpler task to a more complex one, as well as the effects the reward function can have on the robot's behaviour and performance. Also, the popular D3QN algorithm is compared with a more recent variant of the DQN family of methods, the Rainbow algorithm.

\begin{table*}[h!]
\begin{center}
\caption{A non-extensive summary of previous works. There is a set of commonly used RL algorithms, but depending on the choice of state representation, different tasks can be trained.}
\begin{tabular}{p{0.15\textwidth}p{0.15\textwidth}p{0.3\textwidth}p{0.3\textwidth}} 
 \toprule
  Agent & Algorithm & State & Task \\ 
  \midrule
  \cite{indoordqn} & DQN & Depth Image & Obstacle Avoidance \\ [12pt]
  \cite{monoculard3qn} & D3QN with CNN & Predicted Depth Image from RGB &  Obstacle Avoidance \\ [12pt]
  \cite{visionbasedobstacledrl} & PPO and DQN with GAN & Predicted Depth Image from RGB & Maze Navigation \\
 \cite{navd3qn} & D3QN & Depth Image and Distance to Goal & Goal Navigation\\ [12pt]
 \cite{thornav} & A3C & Perspective RGB Image and RGB Image from target & Goal Navigation  \\ [12pt]
  \cite{crowdsarl} & LSTM-RL & Position, Velocity and Radius of Agent and Humans & Goal Navigation in a Crowd \\ [12pt]
  \cite{usingrgb} & DQN with VAE & RGB Image, Polar Coordinates and Motion Information & Goal Navigation \\ [12pt]
  \cite{goobstacle} & DDPG & Depth Images and Polar Coordinates & Goal Navigation \\ [12pt]
  \cite{follownet} & DQN with RNN & Natural Language Instruction, Visual and Depth Data & Goal Navigation with Natural Language Directions \\ [12pt]
  \cite{d3qnslam} & D3QN & Laser Measurements & Active SLAM \\ [12pt]
 \cite{14trackingrl} & A3C with ConvLSTM & RGB Image & Object Following and Tracking\\ [12pt]
  Proposed Agents & D3QN \& Rainbow DQN & Depth Image & Obstacle Avoidance\\[12pt]
   &  & Depth Image and Polar Coordinates & Goal Navigation  \\[12pt]
 \bottomrule
\end{tabular}
\label{alternaterl}
\end{center}
\end{table*}%

For a more in-depth review of DRL algorithms and applications in navigation, the surveys of Zhu et al. \cite{drlreview} or Ye et al. \cite{drlalgorithm} can be consulted. It is noteworthy, as also studied by Zhu et al. in \cite{drlreview}, that more often than not DRL applications in navigation require lightweight navigation solutions to be a complete navigation system. As previously discussed, the most common approach to solve the navigation problem is by using a SLAM technique in a map-building-based robotic system. 

In this work, two different approaches to incorporating a DRL agent in a navigation system are explored. The first one is as an obstacle avoidance agent, which can explore an environment with different obstacles and navigate in circuit-like environments. The second is an agent capable of steering towards a goal when given reference information.
The D3QN and Rainbow DQN algorithms are compared to evaluate the difference in results between an algorithm commonly used and its successor. And finally, different reward functions will be implemented in each method to analyse the difference in results and the actions the agents take.

\section{Design of the DRL agent }

\begin{figure}
    \centering
    \includegraphics[width=8cm]{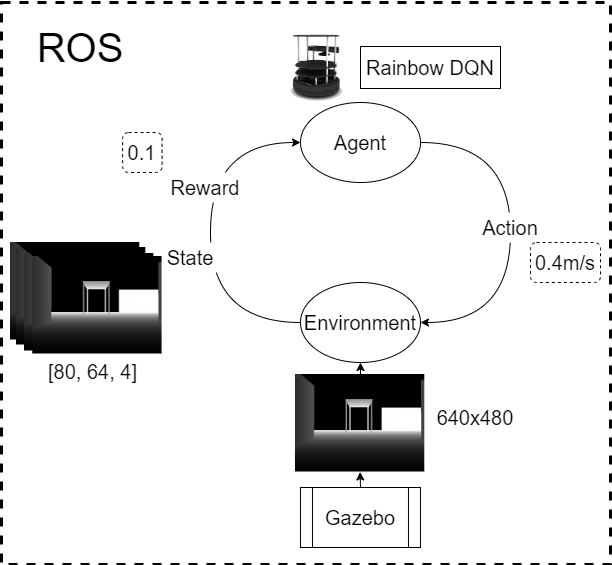}
    \caption{Example of the RL design for the obstacle avoidance task. The depth images are perceived in the simulated environment in Gazebo and reach the RL algorithm through the ROS framework.}
    \label{fig:exampleoa}
\end{figure}

This section contains the details of the DRL approach representation. First, a description of the state representation, action space and reward function will be given. Then, the architecture of the neural networks and specifications of the DRL methods used will be discussed. An intuition of how the implementation for the obstacle avoidance task looks in the agent-environment interaction loop of the RL framework is shown in Fig. \ref{fig:exampleoa}.

\subsection{State Representation}

The state representation must contain enough information about the environment so the agent can decide what action to take to maximise its return, using only the state provided at any given time step.

Depth images provide geometric information about the robot's surroundings in three dimensions. On the contrary, RGB images are more susceptible to lightning and contain colour information which may be irrelevant. Furthermore, depth images are represented in grayscale. This type of image has been proven to be a good state representation in other DRL tasks, such as Atari games \cite{naturedqn}, mainly when used as a stack because of the dense information they contain. For those reasons, the chosen state representation for the obstacle avoidance task consists of a stack of four successive depth images, with one taken at each time step of the training process. The geometric information provided should be enough for the agent to determine when a collision is imminent, and a change of behaviour is necessary.

However, more information is needed to determine the agent's relative position to its destination for the goal-oriented navigation task. To avoid the problem of the agent not recognising the difference between similar states, known as aliasing, the polar coordinates from the agent to the goal are appended to the state representation in the form of a distance and angle.

\subsection{Action Space}

Actions represent the agent's choices to interact with its environment and are constrained by its physical limitations and task. Actions in robotics include desired velocities, accelerations, or torques sent to a motor.

In the case of a mobile robot performing the task of obstacle avoidance, the noteworthy commands are the input linear and angular velocities.
Because the environments are static, there is no action given for the robot to stay still, and it must always remain in motion.
In the case of the discrete set of actions, two linear velocities were selected to allow the robot to either slow down while turning or speed up to reach its destination faster. Also, four angular velocities were chosen to let the robot rotate at different rates in each direction and one null-valued angular velocity to go straight. 

Because of the specifications of the robot used in the simulations for training, the Turtlebot 2, which is further discussed in the next chapter, the specific values are the following: $ 0.2 m/s$ or $0.4 m/s $ for the linear velocity and $\frac{\pi}{6} rad/s$, $\frac{\pi}{12} rad/s$, $0 rad/s$, $\frac{-\pi}{12} rad/s$ or $\frac{-\pi} rad/s$ for the angular velocity. 

\subsection{Reward Function}

The reward function reflects the agent's objective and is the core of the learning process; it grades how well the agent behaved at a given time step.

\subsubsection{Obstacle Avoidance}

For an agent attempting to explore its environment while avoiding obstacles, either a penalty for crashing into an obstacle, a small reward at each time step or a sparse reward for completing several steps without colliding may be enough to learn the task at hand. However, it seems that different approaches may incentive certain behaviours. One such constraint is to penalize the robot's angular velocity for prioritizing moving straight and more steadily. For that reason, two different reward functions were tested. 

The first reward function is a simple one that gives a small reward to an agent for each time step that it does not collide with an obstacle and gives a penalty two orders of magnitude higher on collision:

\begin{equation} \label{eq:1}
 R = \left\{ \begin{array}{ll}
            -10 & \quad \textrm{on collision} \\
        0.1 & \quad \textrm{at each time step} \\
        \end{array}   \right. 
\end{equation}

The second reward function, referred to as the behaviour reward function, rewards the agent for its linear velocity and penalizes the angular velocity:

\begin{equation} \label{eq:2}
 R = \left\{ \begin{array}{ll}
            -10 & \quad \textrm{on collision} \\
        v - |w| & \quad \textrm{at each time step} \\
        \end{array}   \right. 
\end{equation}

Where $ v $ is the linear velocity of the robot and $ w $ is the angular velocity, combined with the previously chosen actions, the robot can earn a reward between $ [-0.13, 0.4] $ at each time step, with the penalty for colliding being two orders of magnitude higher as well, giving it a higher priority when learning.

\subsubsection{Goal-Oriented Navigation}

When the task is changed to a goal-oriented navigation, more information is needed for the agent to receive a reward signal that differentiates whether it is in a better position regarding the goal. For that reason, the chosen metrics were the distance to the goal $ d $ and the heading towards the goal $ \theta $, as the minimum amount of information needed to locate the position of the goal. Thus, the reward function is extended to account for the new information:

\begin{equation} \label{eq:3}
 R = \left\{ \begin{array}{ll}
            -10 & \quad \textrm{on collision} \\
        (v - c|w|) cos(\theta)-v_{max} & \quad \textrm{at each time step} \\
        10 & \quad \textrm{on arrival} \\
        \end{array}   \right. \\
\end{equation}

Where $ cos(\theta) $ determines whether the robot faces the objective and gives a negative reward when the agent strays away, $ c$ is a constant discount factor to avoid the difference between the values of the velocities yields a negative reward, and $v_{max}$ is the maximum linear velocity the robot can achieve. Combined with the previous reward function elements, $v$ and $w$, the agent avoids further penalty when moving straight to the goal and receives more when moving away from it. By increasing the order of magnitude of the reward when reaching the goal, the agent can risk some reward as long as it reaches it. This reward function is referred to as the negative reward function because the values it provides at each step are between $[-0.8,0]$

A positive version of the reward function, where there is no constant penalty based on the maximum linear velocity of the agent, was also used to evaluate which version has better results:

\begin{equation} \label{eq:4}
 R = \left\{ \begin{array}{ll}
            -10 & \quad \textrm{on collision} \\
        (v - c|w|) cos(\theta)& \quad \textrm{at each time step} \\
        10 & \quad \textrm{on arrival} \\
        \end{array}   \right. \\
\end{equation}

Different approaches could have been taken when designing the reward function for such a task, but the current design was chosen, and a sparse reward system was avoided altogether in an attempt for it to generalize and perform better in different kinds of environments.

\subsection{Neural Network Architectures}

A CNN architecture based on the work proposed by Wang et al. \cite{duelingdqn} is used for the D3QN agent, to process the stack of depth images corresponding to the state, and outputting the q-values of each action. The number of layers and hyperparameters of each layer is the same as the NN evaluated in the article. For the case of the goal-oriented navigation task, the distance and angle towards the target are appended to the output of the flattening layer.

For the Rainbow DQN agent, the last layers of the network architecture are modified, following the implementation of the C51 agent described by Bellemare et al. in \cite{distributionalrl}, which uses 51 atoms to estimate the distribution of the rewards instead of the expected values. Training a robotics RL agent in the real world requires a significant amount of time for the algorithm to converge, constant supervision to reset the agent to its initial state after reaching a terminal state, and avoiding accidents. For that reason, the implementation proposed in this thesis is done in a simulator, which has benefits such as speeding up the training time, automatically resetting the whole environment after each episode and allowing different initial configurations for the agent to explore the entire environment better. 

\subsection{Simulated Environment}

The Robotic Operating System (ROS) \footnote{http://wiki.ros.org/} was chosen as the robotics framework to run the experiments, as it provides many software libraries and tools used to build robot applications, as well as communication between the different software needed to run or simulate a robotic system, such as sensor readings, control algorithms and task algorithms. The distribution of ROS used to run the experiments was Melodic Morenia.

The simulated robot used for training is the Turtlebot2 \footnote{https://www.turtlebot.com/turtlebot2/}, an open-source robot commonly used in robotic research. It features an Asus Xtion PRO LIVE as an RGB-D camera and the differential drive base Kobuki, which has a variety of sensors, such as odometry, gyroscope and a laser sensor. Its maximum translational velocity is 0.7 m/s, and its maximum rotational velocity before performance degradation is 110 deg/s. Being differential wheeled allows it to change its direction without additional forward or backward motion. The laser sensor was used to detect collisions at fixed distances accurately. Still, its data were not considered in the state representation, meaning that a bumper or other collision-detecting sensor could replace it.

Gazebo \footnote{https://gazebosim.org/home} was used as the robotics simulator to model the environment, load the Turtlebot2 and its sensors to train the proposed reinforcement learning agent, and speed up the simulation ten times faster than in real-time. The Gazebo version used is 9.0.0. The open-source openai\_ros ROS package, developed by The Construct \footnote{https://www.theconstructsim.com/}, was used as the RL framework, which provides communication between ROS, Gazebo and the RL scripts. It also allows the environment's set-up in Gazebo, which offers states and rewards at each time step and resets the environment at the end of each episode.

Finally, the reinforcement learning algorithms, training and evaluating scripts were implemented using the Python programming language, with the OpenCV computer vision library being used to preprocess the depth images. The Rainbow DQN and D3QN algorithms were based on the implementation of Dittert \cite{atariagents} and Arulkumaran \cite{rainbowpytorch}.

\subsection{Training}

The training was done in a simulated environment. The hyperparameters' values were chosen based on the algorithms' original work. The learning rate, Adam optimiser, gamma, batch size and hidden layer size were the same as the original DQN work of Mnih et al. in \cite{naturedqn}. The buffer size was lowered because of initial hardware limitations, and the number of random steps to fill it was also proportionally decreased. The $N$ step, $\tau$, and minimum $\epsilon$ values were chosen according to the Rainbow DQN proposed by Hessel et al. \cite{rainbowdqn}. The D3QN agent requires the $\epsilon$ hyperparameter for exploration, which starts with a value of 1 and is exponentially decayed until it reaches $\epsilon_{min}$. Since the number of training episodes would be much smaller, compared to other RL-related works, the $\alpha$ value was slightly increased, and the $\omega$ value decreased to prioritise experiences earlier. A soft update of parameters with the value of $\tau$, as described by Lillicrap et al. in \cite{ddpg}, was chosen instead of a hard update. A summary of the hyperparameters used can be seen in the Table \ref{hyperparameters}.

The depth images were resized, normalised and pre-processed before being passed to the agent as observations. The default size of the depth images used for training was $80\times64$ pixels, similar to the size of images used for training RL agents in Atari games since the DQN implementation in \cite{naturedqn}, but keeping the width and height ratio of the original image size. Also, at each step, the depth image was stacked with the three previous ones, as described in the design section, while at the start of each episode, the initial frame was copied four times.

\begin{table}[h!]
\begin{center}
\caption{Hyperparameters values. The D3QN agent requires the hyperparameter $\epsilon$ for exploration, while Rainbow DQN uses the NoisyNets for exploration proposed.}
\begin{tabular}{c  c  c} 
\toprule
 Hyperparameter & D3QN  & Rainbow DQN\\ 
 \midrule
 Learning rate & 0.00025  & 0.00025 \\
 Batch Size & 32 & 32 \\
  Hidden Layer Size & 512 & 512 \\
 $\gamma$ & 0.99 & 0.99 \\
 Buffer Size & 100000 & 100000  \\
 Initial Random Steps & 20000 & 20000  \\
 $\tau$ & 0.001 & 0.001 \\
 $\epsilon_{min}$ & 0.01 & N/A  \\
 N step & 1 & 3 \\
 $\omega$ & 0.4 & 0.4 \\
 $\alpha$ & 0.6 & 0.6 \\
 \bottomrule
\end{tabular}
\label{hyperparameters}
\end{center}
\end{table}%

The experiments were performed on a computer equipped with an AMD Ryzen 5 3600 CPU, an NVIDIA RTX 3060 Ti GPU and 32 GB of RAM.

\subsection{Obstacle Avoidance}

\begin{figure}
     \centering
     \begin{subfigure}[b]{0.49\textwidth}
         \centering
        \includegraphics[width=\textwidth]{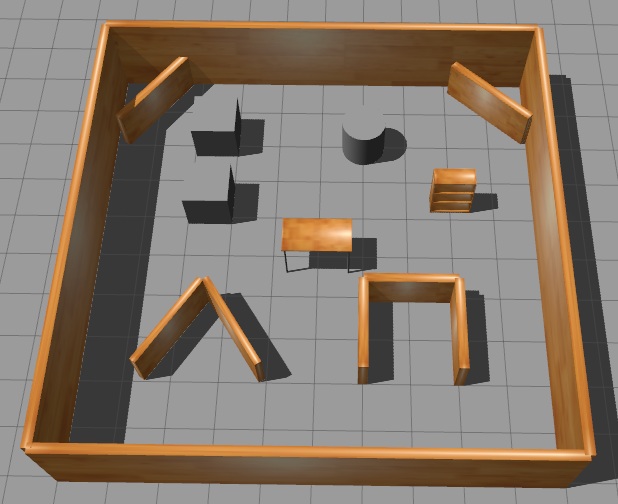}
        \caption{Obstacle Avoidance Environment Perspective View}
        \label{fig:obs_train_env}
     \end{subfigure}
     \hfill
     \begin{subfigure}[b]{0.49\textwidth}
         \centering
        \includegraphics[width=\textwidth]{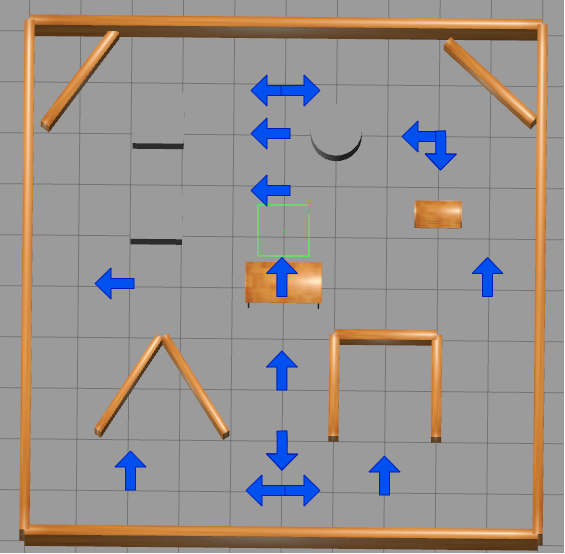}
        \caption{Obstacle Avoidance Training Starting Positions}
        \label{fig:obs_train_env_points}
     \end{subfigure}
        \caption{The training environment for the obstacle avoidance task. Arrows indicate available random starting configurations.}
        \label{fig:obs_train}
\end{figure}

The obstacle avoidance agent was trained in a $ 5m $ environment with different obstacles, as shown in Fig. \ref{fig:obs_train}. The reasoning behind its design was to expose the RL agent to different obstacle shapes to learn better how to avoid collisions. At the start of each episode, the agent's starting position was randomly initialised from 15 possibilities to accelerate the learning process and address the challenge of generalisation presented in \cite{drlreview}. Each training session lasted for 1500 episodes, and the episodes ended after 400 steps or when the agent crashed into an obstacle. For better accuracy, collisions were detected with the robot's laser sensor at a distance of 0.3 meters.

As shown in Table \ref{obsagents}, six obstacle avoidance agents were trained, with their label referring to the algorithm, reward function and size of depth images used during their training.

\begin{table}[h!]
\begin{center}
\begin{tabular}{c  c  c} 
\toprule
  Agent & Reward Function & Size of Depth Image \\ 
  \midrule
  SimpleD3QN & \ref{eq:1} & $80\times64$  \\
  SimpleRainbow & \ref{eq:1} & $80\times64$  \\ 
  SimpleRainbowL & \ref{eq:1} & $160\times128$\\
  BehaviourD3QN & \ref{eq:2} & $80\times64$\\
  BehaviourRainbow & \ref{eq:2} & $80\times64$\\
 BehaviourRainbowL & \ref{eq:2} & $160\times128$\\ 
 \bottomrule
\end{tabular}
\caption{Agents trained for the obstacle avoidance task. Their names indicate their algorithm, reward function and the size of the depth images used during training.}
\label{obsagents}
\end{center}
\end{table}%

\subsection{Navigation}

\begin{figure}
     \centering
     \begin{subfigure}[b]{0.49\textwidth}
         \centering
        \includegraphics[width=\textwidth]{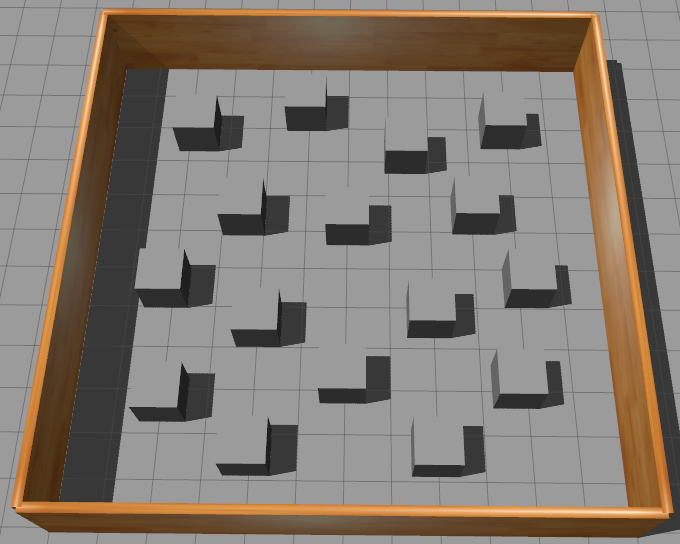}
        \caption{Goal-Oriented Navigation Environment Perspective View}
        \label{fig:nav_train_env}
     \end{subfigure}
     \hfill
     \begin{subfigure}[b]{0.49\textwidth}
         \centering
        \includegraphics[width=\textwidth]{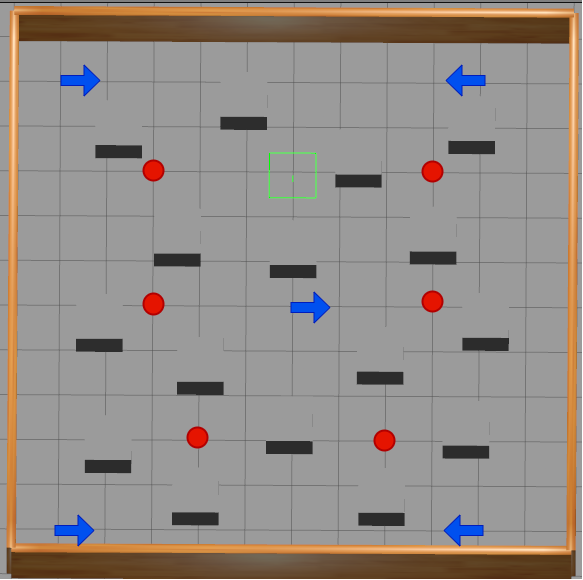}
        \caption{Goal-Oriented Navigation Training Starting and Goal Positions}
        \label{fig:nav_train_env_points}
     \end{subfigure}
        \caption{The training environment for the navigation task. Arrows indicate possible starting configurations, and dots represent goal positions.}
        \label{fig:nav_train}
\end{figure}

The goal-oriented agent was trained in a slightly wider, $6m$ environment with only primitive shapes as obstacles, which can be seen in Fig. \ref{fig:nav_train}. The reason behind using more basic obstacles in the environment is for the agent to focus more on the path-planning competence of the goal-oriented navigation task rather than the obstacle-avoiding one. At the start of each episode, the agent's starting position was randomly initialised from 5 different possibilities and the goal position from a set of 6 cases. The maximum number of steps was slightly lowered to 350, but the total episode count increased to 25,000. The collision detection and episode-ending conditions were almost identical to the previous task,  with an added terminal state when the agent reached the goal. A goal was considered to be contacted at a lenient distance of 0.8$m$ to speed up the learning process.

Three agents were trained, with different algorithm and reward function choices, as seen in Table \ref{navagents}.

\begin{table}[h!]
\begin{center}
\caption{Agents trained for the goal-oriented navigation task. Their names indicate their algorithm and reward function used during training.}
\begin{tabular}{c  c } 
  \toprule
  Agent & Reward Function\\ 
  \midrule
  NegativeD3QN & \ref{eq:3}  \\ 
  NegativeRainbow & \ref{eq:3}  \\ 
  PositiveRainbow & \ref{eq:4}\\ 
  \bottomrule
\end{tabular}
\label{navagents}
\end{center}
\end{table}

\subsection{Evaluation}

\subsubsection{Obstacle Avoidance}

For the obstacle avoidance task, the models were subjected to two evaluations, one for their ability to evade different obstacles and another to test whether their training was enough to navigate a circuit-like environment without a goal. 

\begin{figure}
    \centering
    \includegraphics[width=0.4\textwidth]{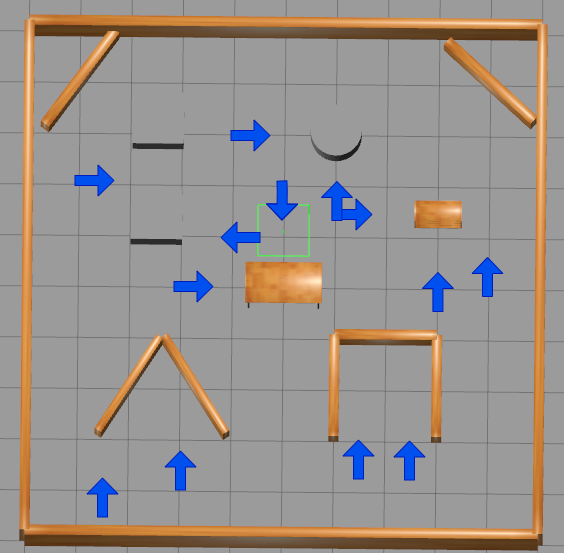}
    \caption{Obstacle avoidance evaluation starting positions. Two points near each obstacle were chosen as valid starting positions.}
    \label{fig:obs_eval2_env}
\end{figure}

The environment used to test obstacle avoidance competence is the same for training but with different starting points that put the robot close to the obstacles from the beginning. Two points nearby were chosen as starting positions for each of the six types of obstacles, resulting in 12 initial configurations, as seen in Fig. \ref{fig:obs_eval2_env}. The evaluation had a duration of 600 episodes, with 100 steps each. The idea behind it is only to check whether the robot can avoid collisions with the specific types of obstacles it is trained with, as its capability to move forward while avoiding walls will be tested later. For the obstacle avoidance task, the models were subjected to two evaluations, one for their ability to evade different obstacles and another to test whether their training was enough to navigate a circuit-like environment without a goal. 

\begin{figure}
     \centering
     \begin{subfigure}[b]{0.49\textwidth}
         \centering
        \includegraphics[width=\textwidth]{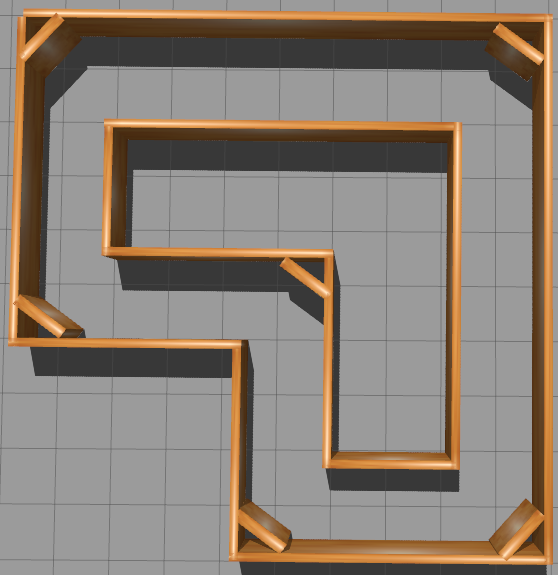}
        \caption{Circuit Navigation Environment Perspective View}
        \label{fig:obs_eval3_env}
     \end{subfigure}
     \hfill
     \begin{subfigure}[b]{0.49\textwidth}
         \centering
        \includegraphics[width=\textwidth]{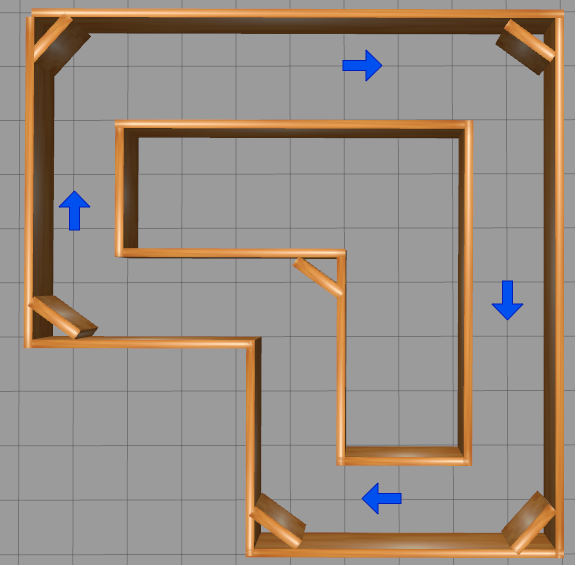}
        \caption{Circuit Navigation Evaluation Starting Positions}
        \label{fig:obs_eval3_env_points}
     \end{subfigure}
        \caption{Circuit navigation evaluation environment for the obstacle avoidance agents. Although the obstacles are simpler, the lack of space prevents circular motion from being an optimal behaviour to avoid collisions.}
        \label{fig:obs_eval}
\end{figure}

The second test was performed in a simple circuit-like environment with four pre-defined starting points, shown in Fig. \ref{fig:obs_eval}. A perfect performance was not expected, as the agent was trained in a different environment. However, the reasoning behind it is that RL agents sometimes optimise their behaviour in unintended ways. One such case for an obstacle avoidance task, as there is no reward based on a clear objective other than a penalty for colliding, would be if the agent moved around in circles. To test whether the agents can navigate a road bounded by walls where circular motion is impossible without colliding, a simple circuit-like scenario from the openai\_ros package was adapted as an evaluation environment.

In both cases, the distance for considering a collision was slightly lowered to $0.2m$ to evaluate the agent's reaction competence better.

\subsubsection{Navigation}

\begin{figure}
    \centering
    \includegraphics[width=0.4\textwidth]{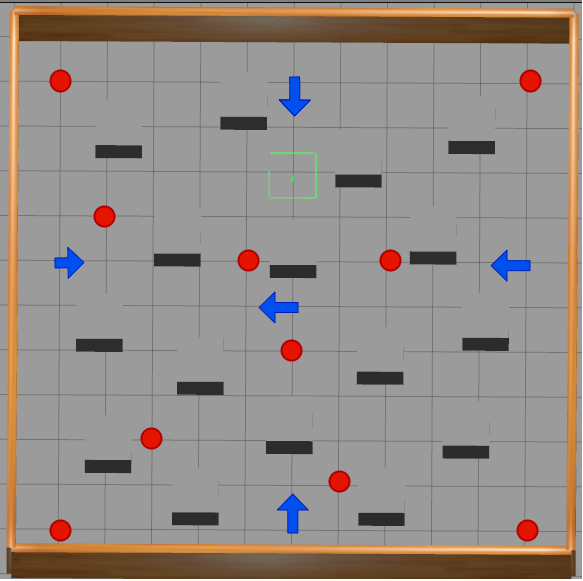}
    \caption{Goal-Oriented navigation evaluation environment. The starting and goal positions were shifted to test the agents' adaptation capacity.}
    \label{fig:nav_eval_env}
\end{figure}

For the navigation task, the models were evaluated in the same environment used for training, with and without the same starting points. The assignment was more challenging, as the agents needed to avoid obstacles while moving closer to the goal. Therefore, each agent was tested on its learning and adaptative capabilities. The agent was allowed to navigate a maximum of 250 steps to reach its destination and was evaluated for 1000 episodes. The number of goal positions was increased to 10, but the starting configurations were kept to 5. The adjustment of starting and goal positions is shown in Fig. \ref{fig:nav_eval_env}

The collision detection was turned off during the evaluation process so that the agent still had a chance to overcome the obstacles and fulfil the goal-reaching task.


\section{Experimental Results}

\subsection{Training}

There are different metrics to consider when evaluating the training performance of an RL agent. The most important is the return, which indicates how well the agent performed its task. However, in the goal-oriented navigation task, the starting and goal positions are randomly chosen from a set at the start of each episode, meaning that the maximum return the agent can achieve per episode varies; therefore, the metric can be pretty noisy. Nonetheless, it still shows the learning curve and is expected to increase over time as the agent optimises its behaviour. One task-independent metric, also used in ML applications, to describe the learning of an algorithm is the loss function, which is expected to decrease over time as the agent explores its environment and improves its estimations. The loss indicates the mean squared error between the $q$ value calculated and the expected value for the TD methods.
A Task dependent metric that can be compared for the navigation task is the percentage of times the agent reaches the goal. As for the obstacle avoidance task, the rate of collisions and steps the agent managed to navigate before crashing can be measured. 

Because the original plots are very noisy, mainly due to the initial random position at the start of each episode, the results presented were calculated using a moving average of one hundred steps.

Finally, the different metrics were measured in episodes, as the tasks relied on avoiding collisions or reaching the goal within a reasonable amount of time steps, and the agents were rewarded or punished accordingly. The only exception was the loss function, which was monitored at each time step to verify the learning process with each batch of samples used.

\subsubsection{Obstacle Avoidance}

Six different agents were trained and compared for the obstacle avoidance task, three of which consist of a D3QN agent and two Rainbow agents trained with varying sizes of depth images, using the simple reward function. 

\begin{figure}
     \centering
     \begin{subfigure}[b]{0.49\textwidth}
         \centering
        \includegraphics[width=\textwidth]{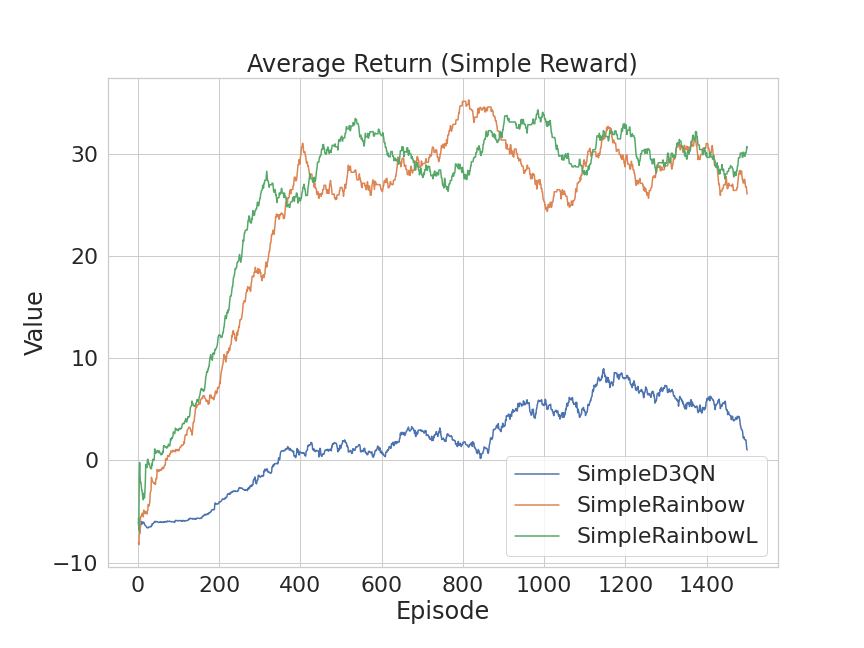}
        \caption{Return Comparison}
        \label{fig:returnobsv1}
     \end{subfigure}
     \hfill
     \begin{subfigure}[b]{0.49\textwidth}
         \centering
        \includegraphics[width=\textwidth]{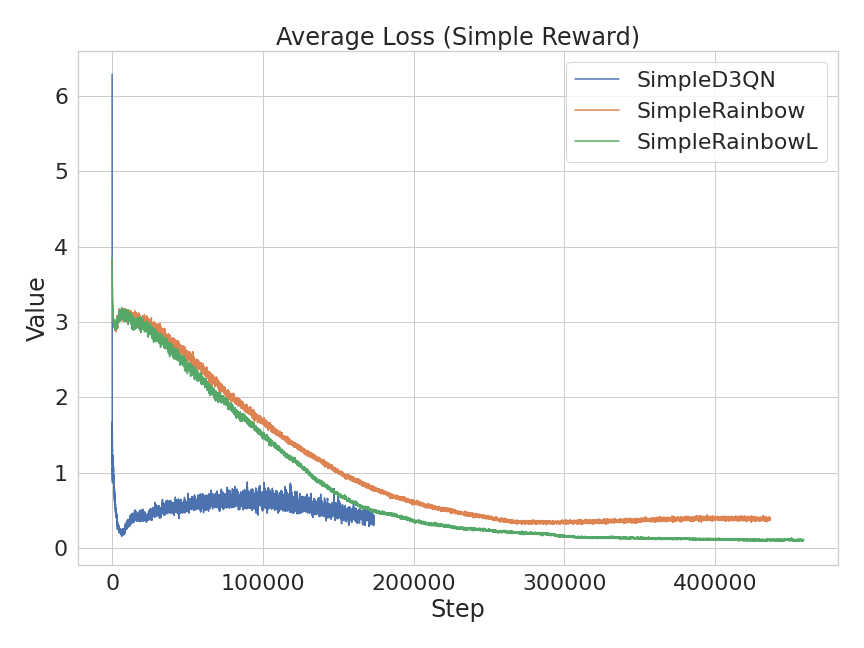}
        \caption{Loss Comparison}
        \label{fig:lossobsv1}
     \end{subfigure}
     \begin{subfigure}[b]{0.49\textwidth}
         \centering
        \includegraphics[width=\textwidth]{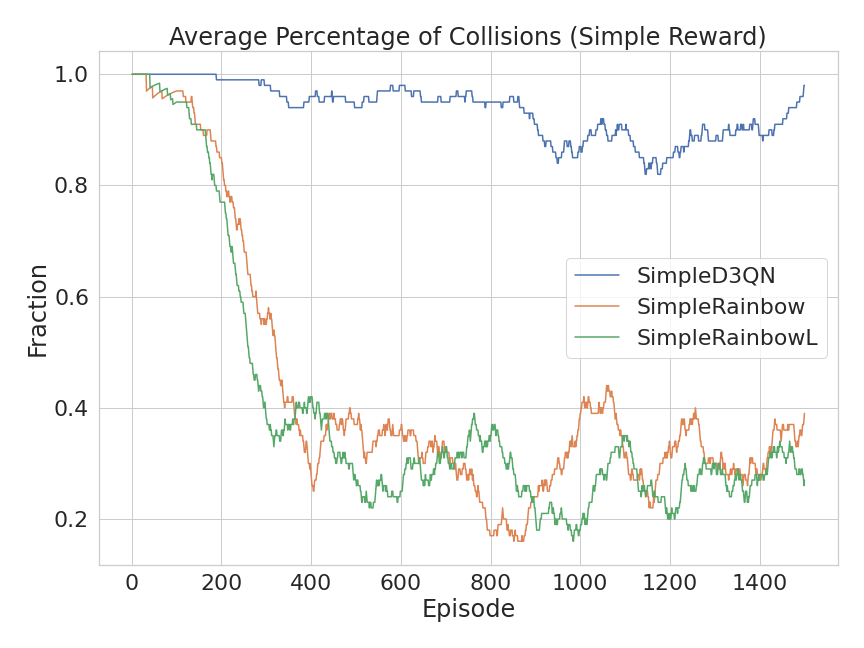}
        \caption{Collision Comparison}
        \label{fig:crashobsv1}
     \end{subfigure}
     \hfill
     \begin{subfigure}[b]{0.49\textwidth}
         \centering
        \includegraphics[width=\textwidth]{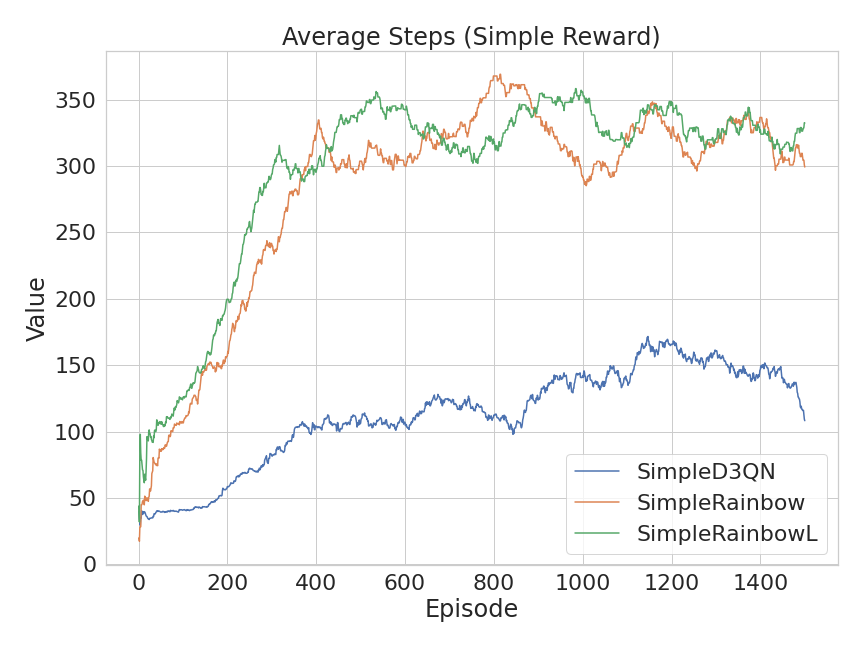}
        \caption{Steps Comparison}
        \label{fig:stepsobsv1}
     \end{subfigure}
        \caption{Training performance of the obstacle avoidance agents with simple reward. The best performance was achieved by SimpleRainbowL, the Rainbow DQN agent that used the simple reward function and larger depth image size.}
        \label{fig:obstrainv1}
\end{figure}

Between the agents with the simple reward function, which corresponds to the equation \ref{eq:1}, SimpleRainbowL achieved slightly better results than SimpleRainbow by maintaining a higher return, lower collision rate and more training steps, as seen in Fig. \ref{fig:obstrainv1}. The D3QN was trained for fewer training steps, meaning it crashed earlier in each episode. The use of a smaller depth image size allowed SimpleRainbow to seemingly achieve peak performance at around 800 episodes, followed by SimpleRainbowL at 1000 episodes and SimpleD3QN at 1200 episodes, when their amount of return was at its highest and collision rate at its lowest.

\begin{figure}
     \centering
     \begin{subfigure}[b]{0.49\textwidth}
         \centering
        \includegraphics[width=\textwidth]{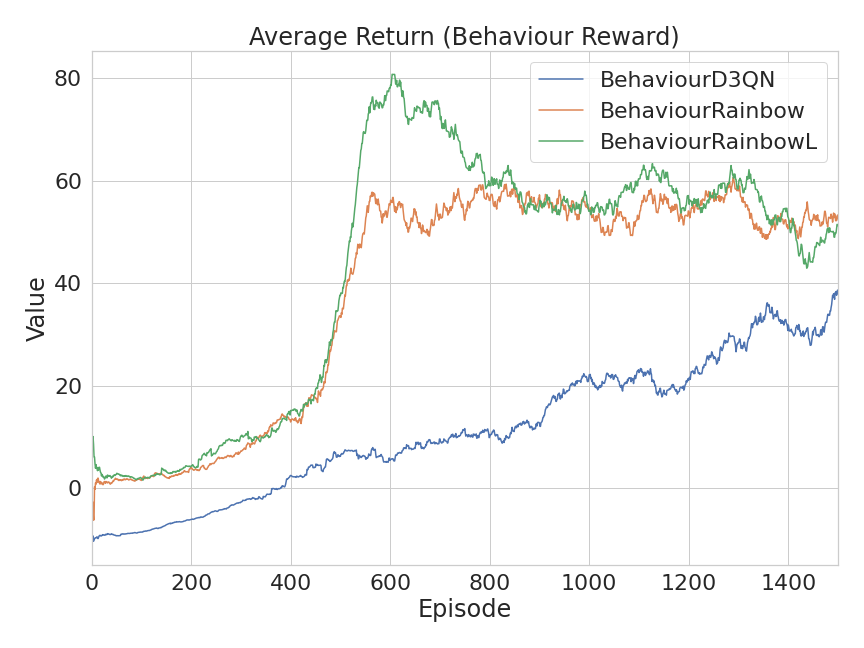}
        \caption{Return Comparison}
        \label{fig:returnobsv2}
     \end{subfigure}
     \hfill
     \begin{subfigure}[b]{0.49\textwidth}
         \centering
        \includegraphics[width=\textwidth]{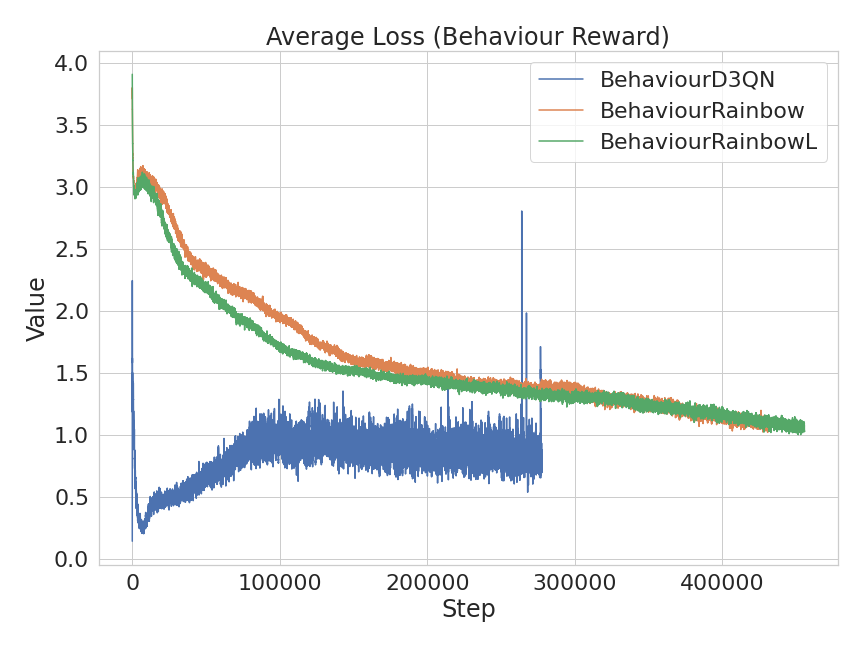}
        \caption{Loss Comparison}
        \label{fig:lossobsv2}
     \end{subfigure}
     \begin{subfigure}[b]{0.49\textwidth}
         \centering
        \includegraphics[width=\textwidth]{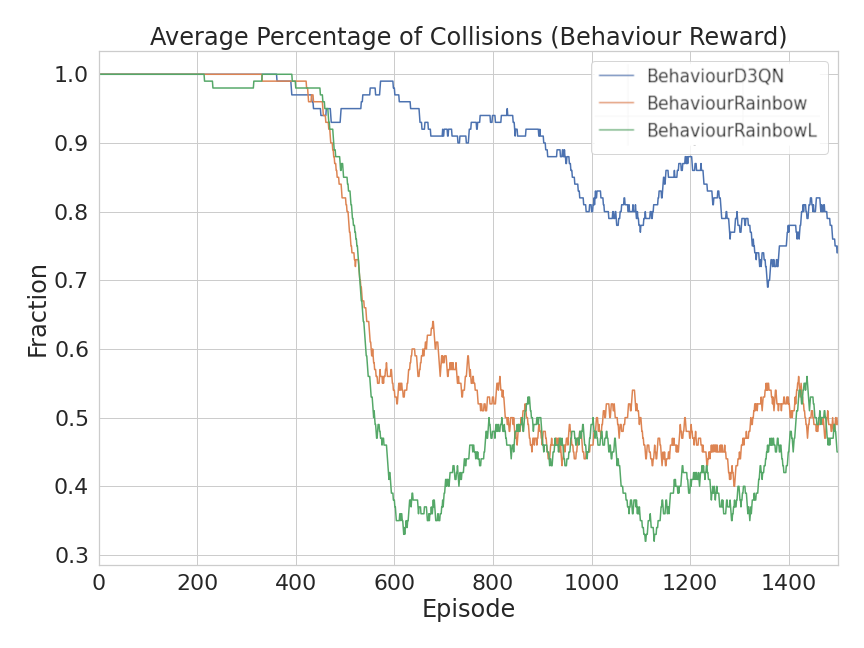}
        \caption{Collision Comparison}
        \label{fig:crashobsv2}
     \end{subfigure}
     \hfill
     \begin{subfigure}[b]{0.49\textwidth}
         \centering
        \includegraphics[width=\textwidth]{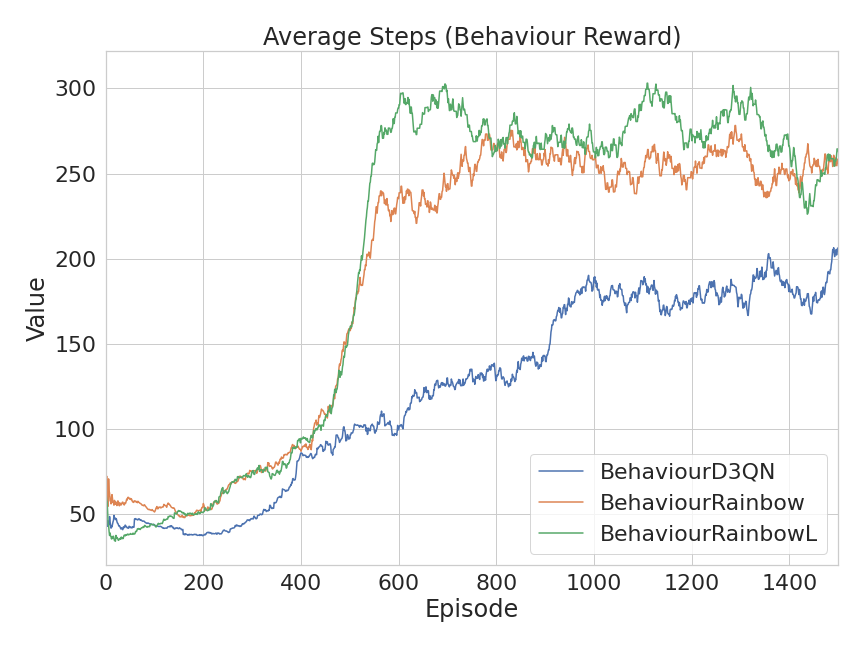}
        \caption{Steps Comparison}
        \label{fig:stepsobsv2}
     \end{subfigure}
        \caption{Training performance of the obstacle avoidance agents with behaviour reward. The best performance was achieved by BehaviourRainbowL, the Rainbow DQN agent that used the behaviour reward function and larger depth image size.}
        \label{fig:obstrainv2}
\end{figure}

Similar results were achieved by the agents with the behavioural reward, corresponding to the equation \ref{eq:1}, as demonstrated in Fig. \ref{fig:obstrainv2}. However, peak performance was achieved after 1100 episodes, indicated by the collision rate, as the return now depends on the behaviour. BehaviourRainbowL had a lower collision rate and higher return, meaning that it performed better at the task and at adapting to the constraints in the velocities.

\begin{figure}
     \centering
     \begin{subfigure}[b]{0.49\textwidth}
         \centering
        \includegraphics[width=\textwidth]{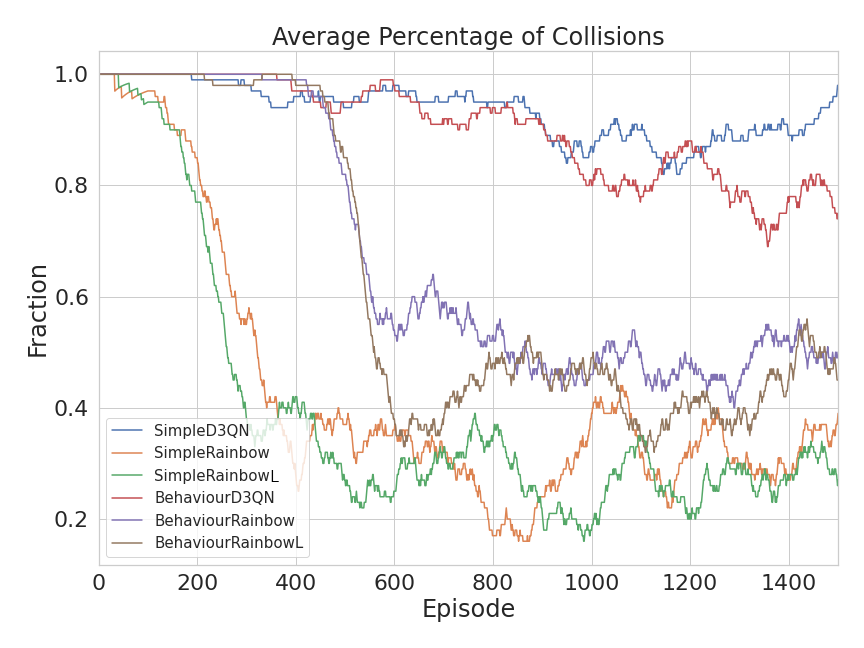}
        \caption{Collision Comparison}
        \label{fig:crashobs}
     \end{subfigure}
     \hfill
     \begin{subfigure}[b]{0.49\textwidth}
         \centering
        \includegraphics[width=\textwidth]{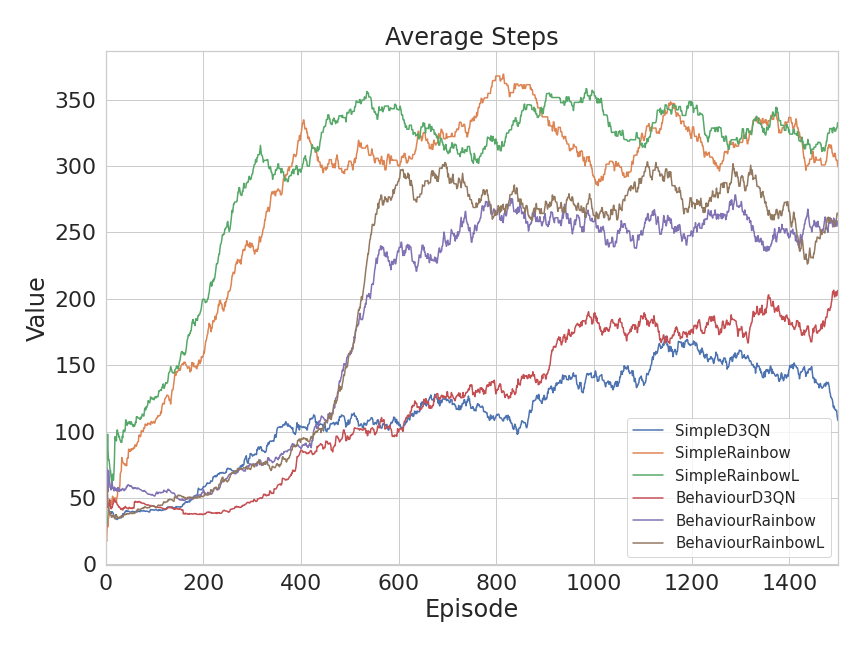}
        \caption{Steps Comparison}
        \label{fig:stepsobs}
     \end{subfigure}
        \caption{Training performance of the obstacle avoidance agents. Better results were achieved by using the Rainbow DQN algorithm, the simple reward function and a larger depth image size.}
        \label{fig:obstrain}
\end{figure}

As seen in Fig. \ref{fig:obstrain}, the Rainbow agents outperformed the D3QN ones at avoiding collisions by doubling the number of steps navigated and having much lower crash rates. Also, using a larger image slightly improved the results, but at the cost of requiring more time to train. The loss and returns could not be compared, as the reward functions operated at different scales. Agents with the behaviour reward took longer to learn to avoid collisions, as they seemed to start optimising their behaviour first. Still, their performance increased sharply after some exploration, which can be seen in the drop of their collision rate in Fig. \ref{fig:obstrain}. Even so, as expected, the agents with the simple reward had lower collision rates, as their only objective was to avoid collisions. In contrast, the behaviour reward imposed a penalty on the other agents' choice of speed, which demanded more training time to improve their results.

All agents learnt at different rates, as seen in the decrease in their average loss.

\subsubsection{Navigation}

For the navigation task, three different agents were trained, the D3QN agent and a Rainbow agent for each of the two reward functions. In this case, the average return cannot be compared, as both reward functions are on a different scale but can be seen as the agent's learning process. Also, the average amount of steps was not used as a metric, as most of the time, the agents collided quickly, and the episode ended early while they learnt to reach the goal.

\begin{figure}
     \centering
     \begin{subfigure}[b]{0.6\textwidth}
         \centering
        \includegraphics[width=\textwidth]{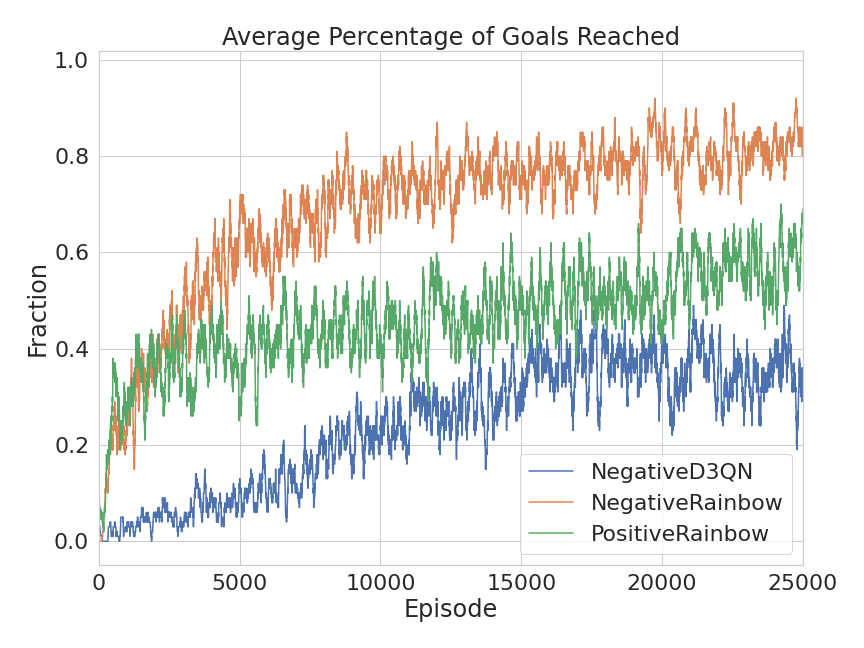}
        \caption{Goals Comparison}
        \label{fig:goalsnav}
     \end{subfigure}
     \begin{subfigure}{0.49\textwidth}
         \centering
        \includegraphics[width=\textwidth]{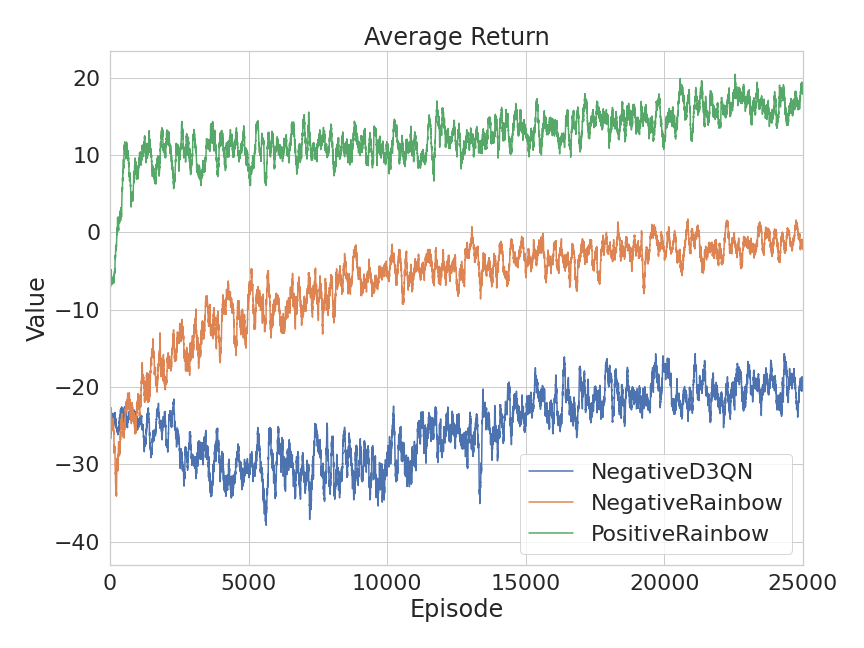}
        \caption{Return Comparison}
        \label{fig:returnnav}
     \end{subfigure}
     \begin{subfigure}{0.49\textwidth}
         \centering
        \includegraphics[width=\textwidth]{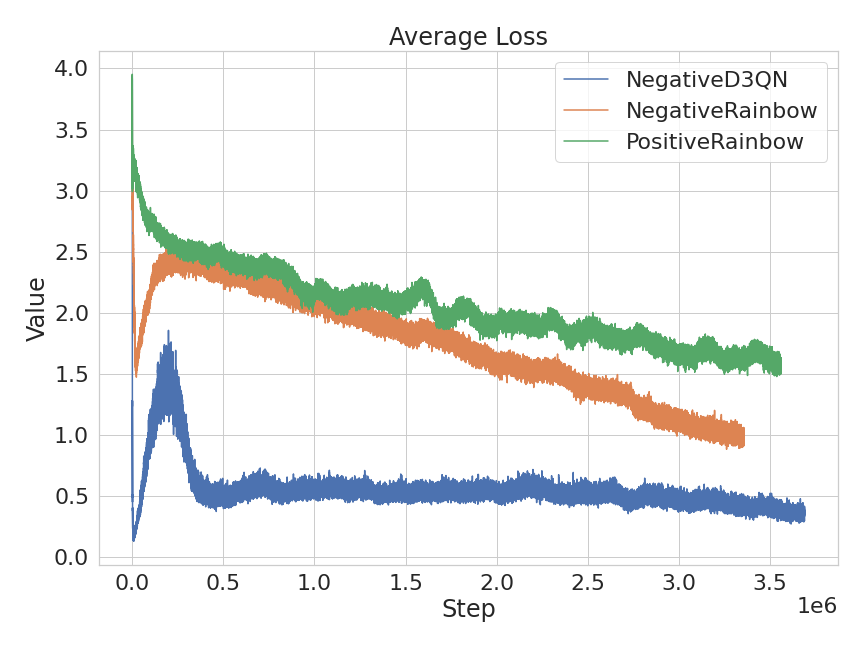}
        \caption{Loss Comparison}
        \label{fig:lossnav}
     \end{subfigure}
        \caption{Training performance of the goal-oriented navigation agents. NegativeRainbow performed the task better by achieving a higher rate of goals reached. Meanwhile, the loss and return evidenced the learning process of all agents.}
        \label{fig:navtrain}
\end{figure}

As evidenced in Fig. \ref{fig:navtrain}, Rainbow agents performed better than the D3QN agent by doubling the number of times they reached the goal. Additionally, NegativeRainbow, the agent with the negative-based reward function corresponding to the equation \ref{eq:3}, yielded better results by reaching the goal more often, as expected. The reason is that the additional constraint encourages the agent to reach the destination as fast as possible to stop the punishment at each time step. Nonetheless, the positive reward-based agent, which follows the equation \ref{eq:4}, still managed to optimise its behaviour and reach the goal a fair amount of times, as seen by its increasing return. 

All the agents had more room to learn, as seen in their increasing returns and decreasing losses at the end of the training process.

\subsection{Evaluation}

For the evaluation process, only the task-dependent metrics are compared, as there is no learning process involved, and the trained models are only used to select their best-valued action, given the current state of the environment. Only Rainbow agents were used for evaluation, as they drastically outperformed the D3QN agents during training and were expected to perform better even under different conditions. 

\subsubsection{Obstacle Avoidance}

For both tests, obstacle avoidance and circuit navigation, the agents were evaluated on their average crash rate and the average number of steps they could navigate without a collision. Also, the action selected at each time step was tracked to analyse the behaviour of each agent.

The results for the evaluation of the obstacle evasion task were similar to those during training. SimpleRainbow achieved the best results, as seen in Table \ref{obsperf}, by having a lower collision rate and a higher number of steps without crashing. Using the simple reward function almost halved the average collision percentage than using the behaviour reward function, and a larger depth image size also produced slightly better results. The average collision rates are higher than the final averages seen during the training process in Fig. \ref{fig:obstrain}, as the difference in conditions influences the results.

\begin{figure}
     \centering
     \begin{subfigure}[b]{0.49\textwidth}
         \centering
        \includegraphics[width=\textwidth]{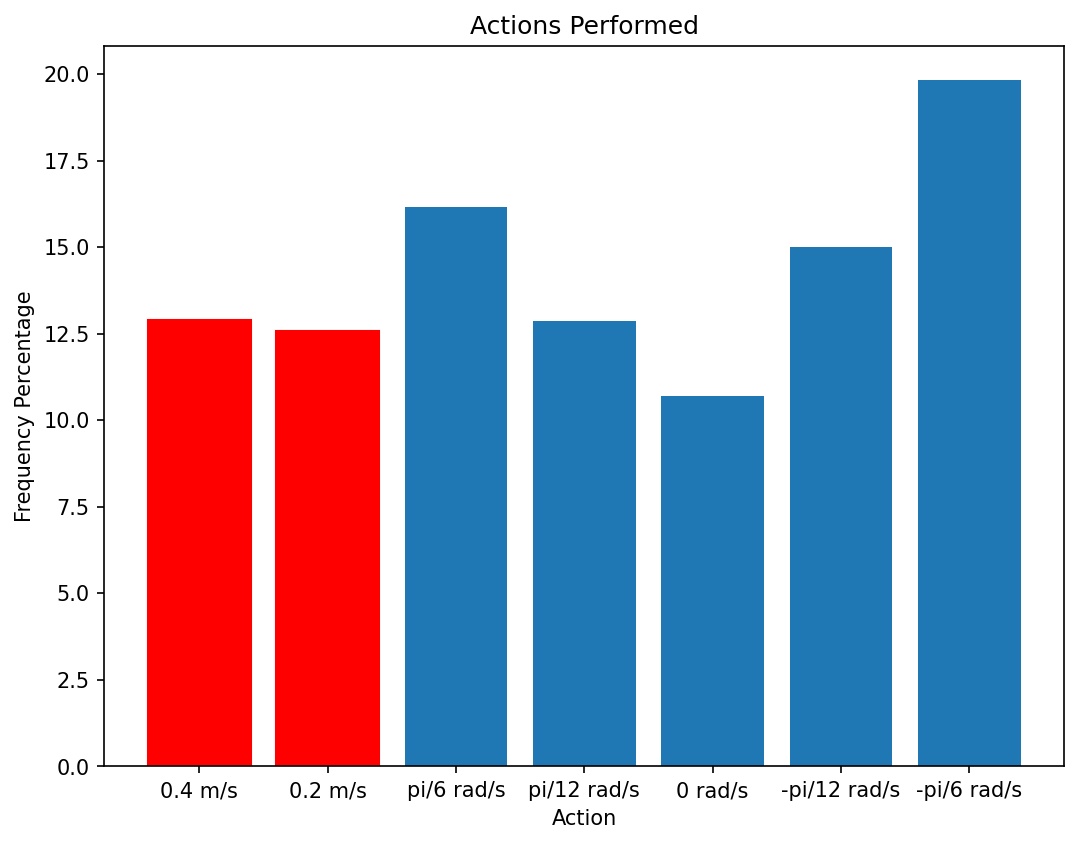}
        \caption{SimpleRainbow}
        \label{fig:act1}
     \end{subfigure}
     \hfill
     \begin{subfigure}[b]{0.49\textwidth}
         \centering
        \includegraphics[width=\textwidth]{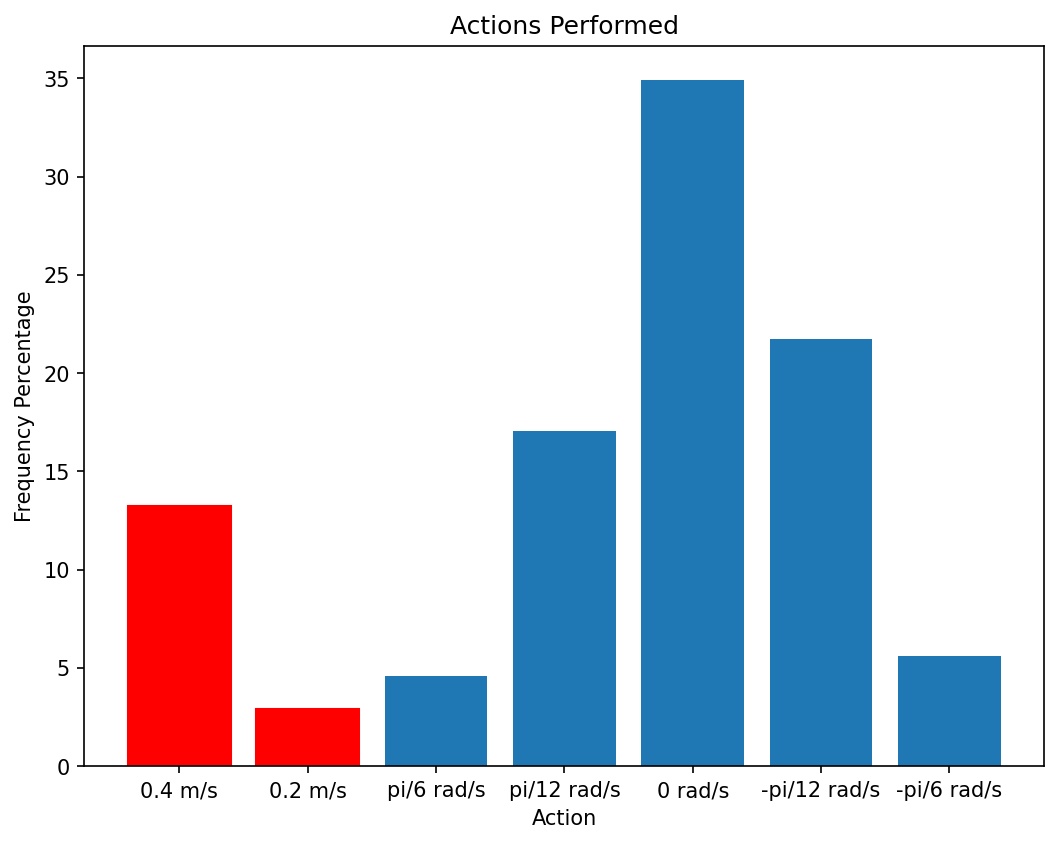}
        \caption{BehaviourRainbow}
        \label{fig:act2}
     \end{subfigure}
     \begin{subfigure}[b]{0.49\textwidth}
         \centering
        \includegraphics[width=\textwidth]{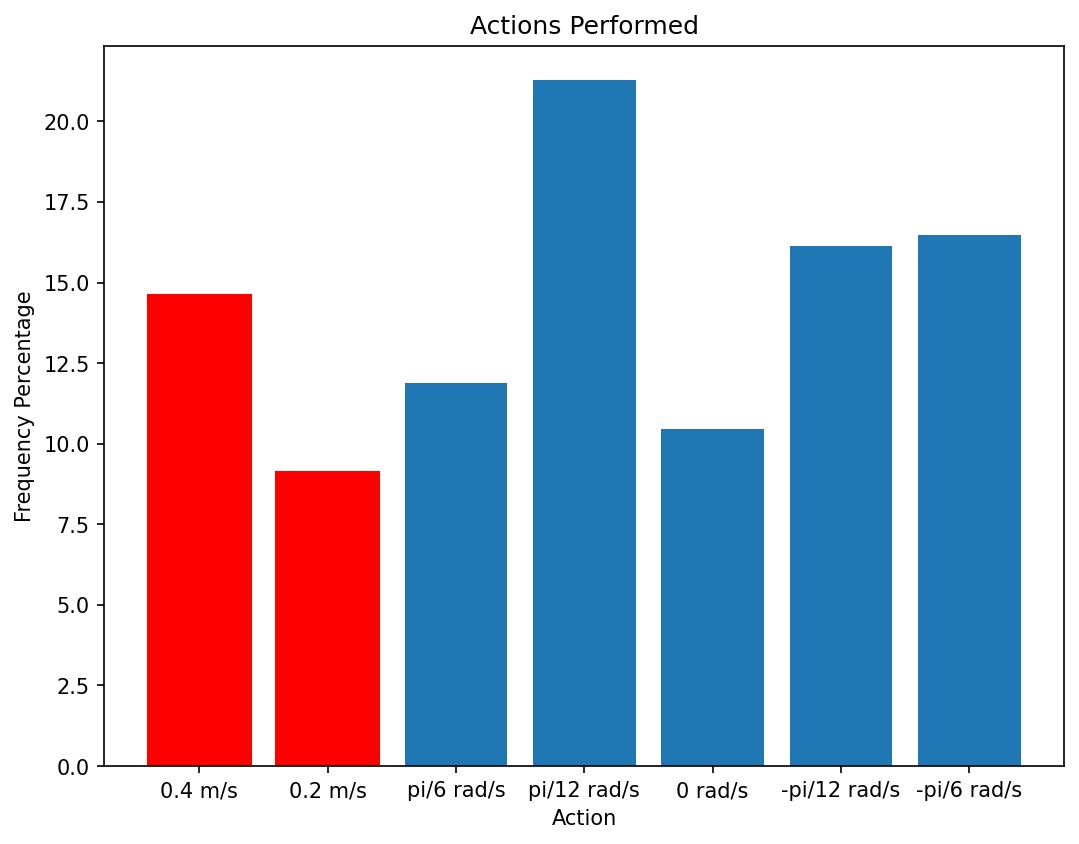}
        \caption{SimpleRainbowL}
        \label{fig:act3}
     \end{subfigure}
     \hfill
     \begin{subfigure}[b]{0.49\textwidth}
         \centering
        \includegraphics[width=\textwidth]{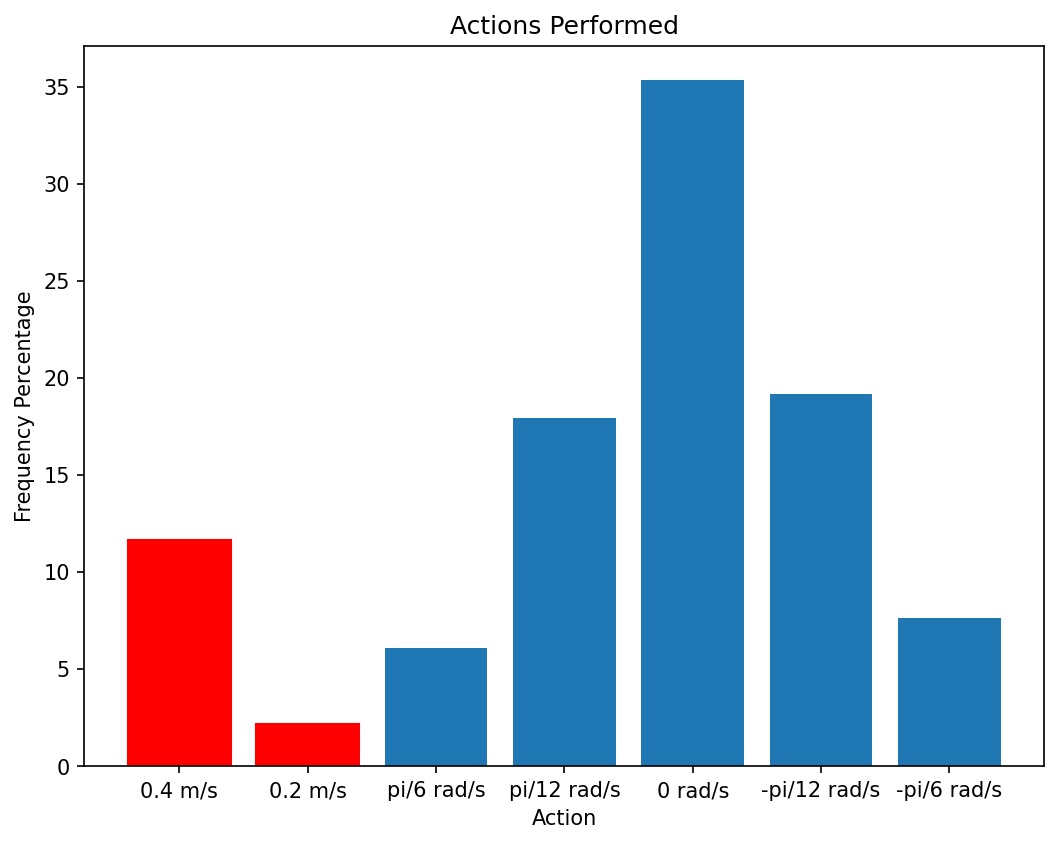}
        \caption{BehaviourRainbowL}
        \label{fig:act4}
     \end{subfigure}
        \caption{Distributions of the chosen actions by the obstacle avoidance agents during evaluation. The behaviour reward function restricted the choice of angular speeds and prioritised the maximum value of linear speed.}
        \label{fig:obsactions}
\end{figure}

However, the difference in performance can be related to the difference in each agent's chosen actions distribution, seen in Fig. \ref{fig:obsactions}. The agents with a simple reward function had a uniform distribution in their action selection, with a slight preference for evading a particular direction. Meanwhile, the agents with the behaviour reward function prioritised using the highest linear velocity and avoiding turning altogether, preferring small angular velocities when it is necessary to avoid an obstacle. For that, the difference in task performance is unsurprising when considering that one type of agent had to evade while going at full speed and barely turning.

\begin{table}
\begin{center}
\caption{Evaluation performance of the obstacle avoidance agents. SimpleRainbowL achieved the best results by colliding less and persisting for more time without crashing.}
\begin{tabular}{c  c  c} 
\toprule
  Agent & Average Collision Percentage & Average Steps \\ 
  \midrule
  SimpleRainbow & 43.16\% & 69  \\ 
  SimpleRainbowL & \textbf{41.83\%} & \textbf{71}\\ 
  BehaviourRainbow & 74.83\% & 45\\
 BehaviourRainbowL & 70.5\% & 47\\ 
 \bottomrule
\end{tabular}
\label{obsperf}
\end{center}
\end{table}

In the case of the circuit navigation evaluation, using a larger depth image size proved to be more critical than the reward function, as seen in Table \ref{mazeperf} that those agents collided less, independently of their reward function. Still, SimpleRainbowL achieved a lower collision rate and a higher number of steps without crashing during training, showing its better ability to adapt to a different environment. Nonetheless, the results were better than expected, with all agents being able to navigate above the average amount of steps and only having difficulties in the sharp turns of the circuit, which were absent in their training environment. In addition, the best agent, SimpleRainbowL, reached below the halfway mark for the average amount of collisions, as seen in Table \ref{mazeperf}, with its evident difficulty being the left turn at the centre of the scene, which can be reached from two out of the four starting points, corresponding to the right and bottom starting positions in Fig. \ref{fig:obs_eval}. 

\begin{figure}
     \centering
     \begin{subfigure}[b]{0.49\textwidth}
         \centering
        \includegraphics[width=\textwidth]{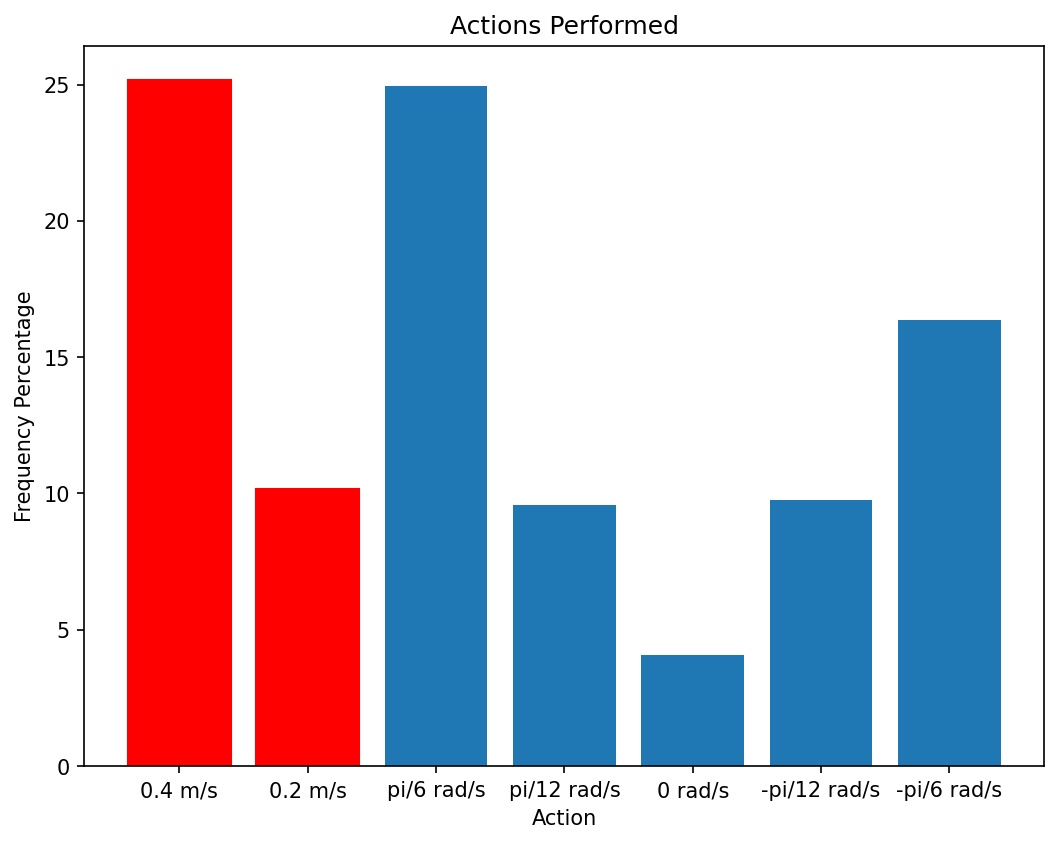}
        \caption{SimpleRainbow}
        \label{fig:act1v2}
     \end{subfigure}
     \hfill
     \begin{subfigure}[b]{0.49\textwidth}
         \centering
        \includegraphics[width=\textwidth]{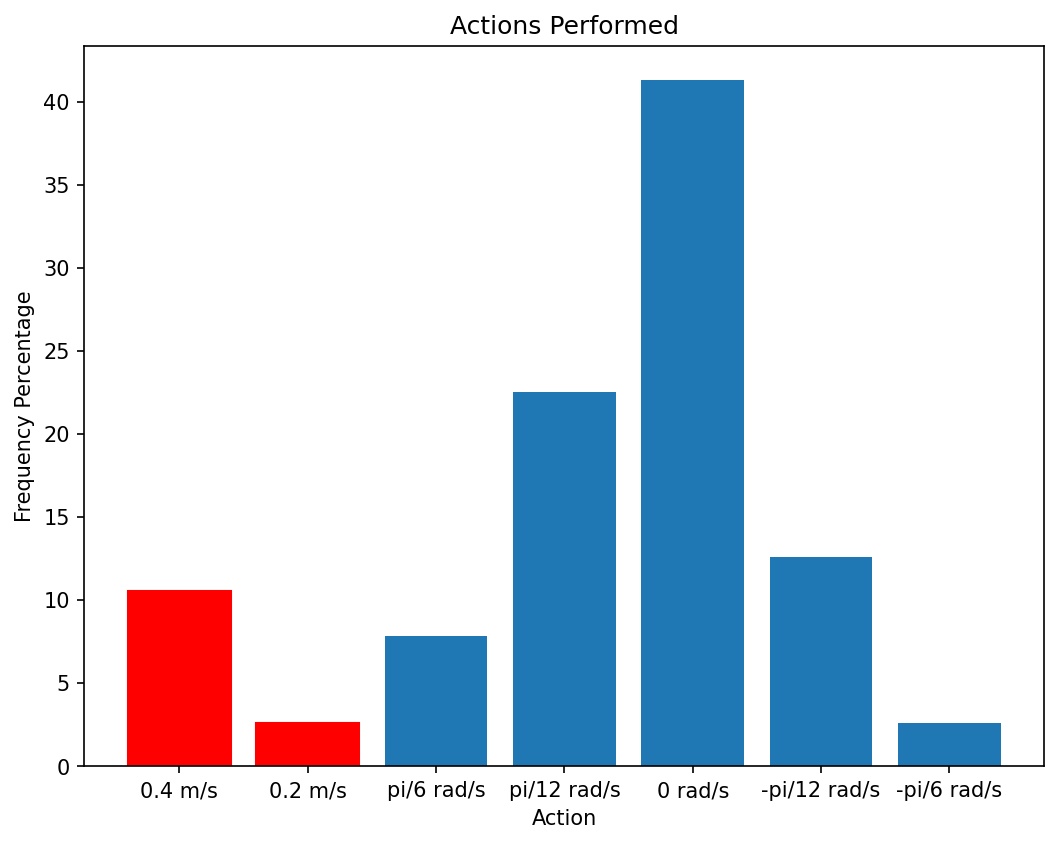}
        \caption{BehaviourRainbow}
        \label{fig:act2v2}
     \end{subfigure}
     \begin{subfigure}[b]{0.49\textwidth}
         \centering
        \includegraphics[width=\textwidth]{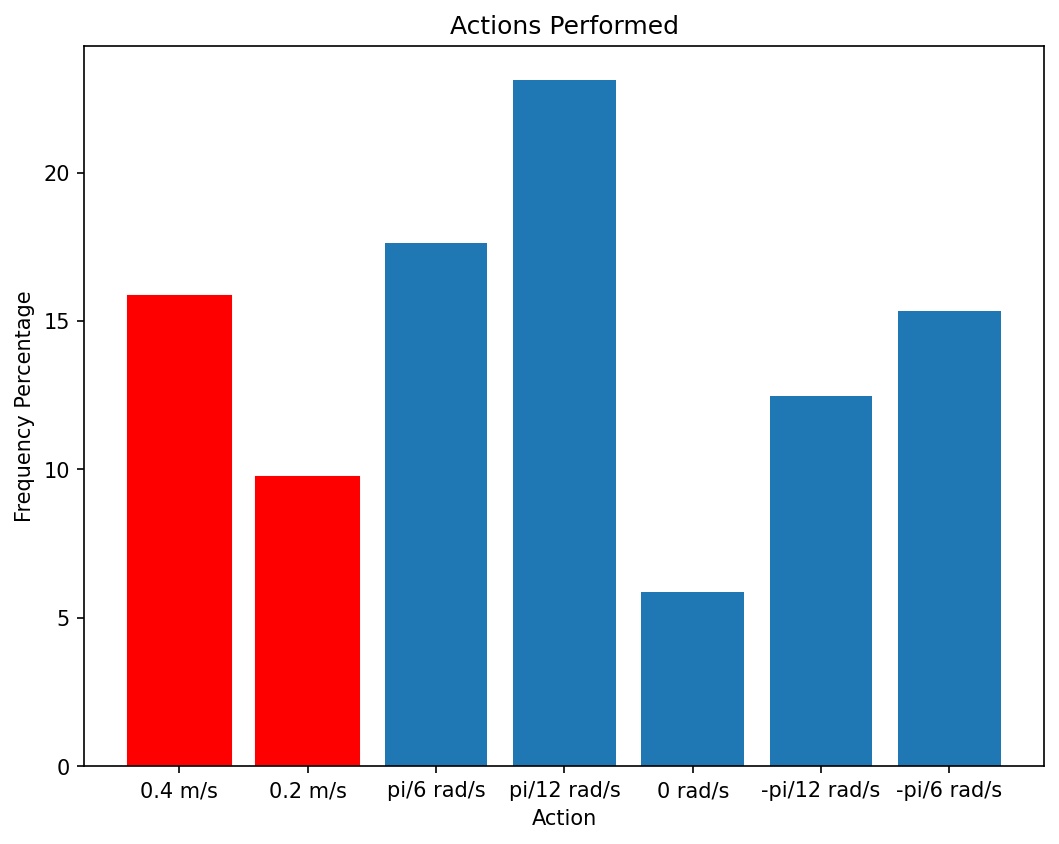}
        \caption{SimpleRainbowL}
        \label{fig:act3v2}
     \end{subfigure}
     \hfill
     \begin{subfigure}[b]{0.49\textwidth}
         \centering
        \includegraphics[width=\textwidth]{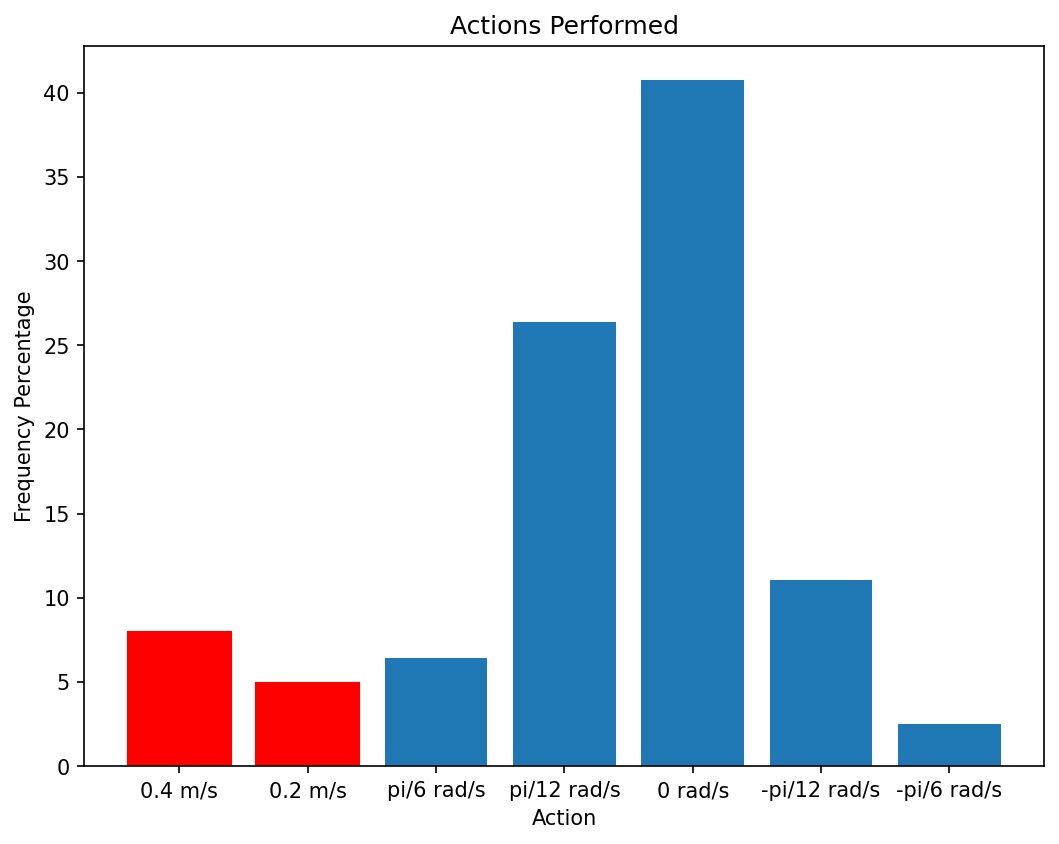}
        \caption{BehaviourRainbowL}
        \label{fig:act4v2}
     \end{subfigure}
        \caption{Distributions of the chosen actions by the obstacle avoidance agents during circuit navigation evaluation. The increase in the need to turn right further evidences the difference in behaviour and performance. The simple reward function allowed the agents to overcome the circuit's sharp turns better.}
        \label{fig:labactions}
\end{figure}

The contrast of the chosen actions distribution is also seen in this task, evidenced by the Fig. \ref{fig:labactions}, with the behaviour reward function demanding less turning and more speed. There was an increase in the choice of turning right, which was caused by the circuit design.

\begin{table}
\begin{center}
\caption{Evaluation performance of the obstacle avoidance agents in the circuit navigation. SimpleRainbowL achieved the best results by colliding less, even in an environment with sharper turns.}
\begin{tabular}{c c c} 
\toprule
  Agent & Average Collision Percentage & Average Steps \\ 
  \midrule
  SimpleRainbow & 72.5\% & 109  \\ 
  SimpleRainbowL & \textbf{48\%} & \textbf{150}\\ 
  BehaviourRainbow & 82.25\% & 115\\
 BehaviourRainbowL & 58.25\% & 146\\ 
 \bottomrule
\end{tabular}
\label{mazeperf}
\end{center}
\end{table}

\subsubsection{Navigation}

The navigation agents were evaluated in the same scene as their training, and the collisions were turned off to test better their learnt ability to reach the goal. When evaluated under the same training conditions, as noticed in Table \ref{navevaltab2}, using the negative-based reward achieves better results, almost beating the environment altogether. Nonetheless, the positive reward-based agent also achieved good results, reaching the goal around seventy per cent of the time. The lower average number of steps of NegativeRainbow proves its speed at reaching its destination.

\begin{figure}
     \centering
     \begin{subfigure}[b]{0.49\textwidth}
         \centering
        \includegraphics[width=\textwidth]{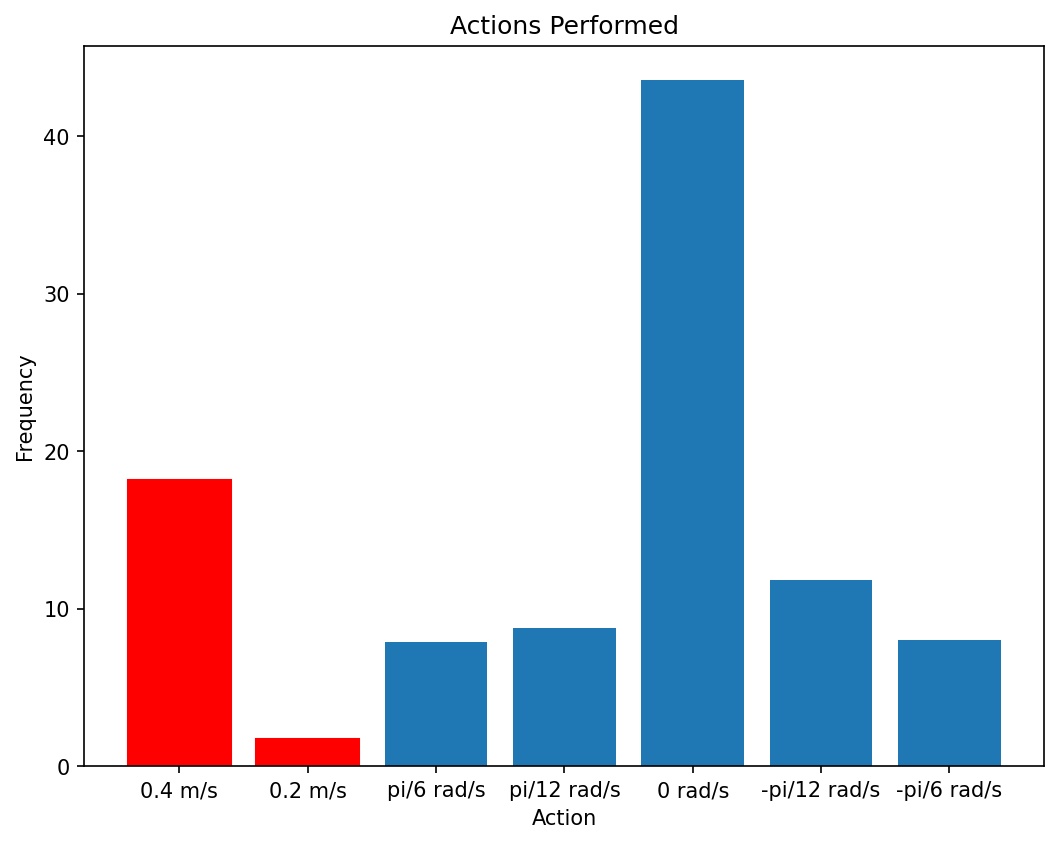}
        \caption{NegativeRainbow}
        \label{fig:navact1v2}
     \end{subfigure}
     \hfill
     \begin{subfigure}[b]{0.49\textwidth}
         \centering
        \includegraphics[width=\textwidth]{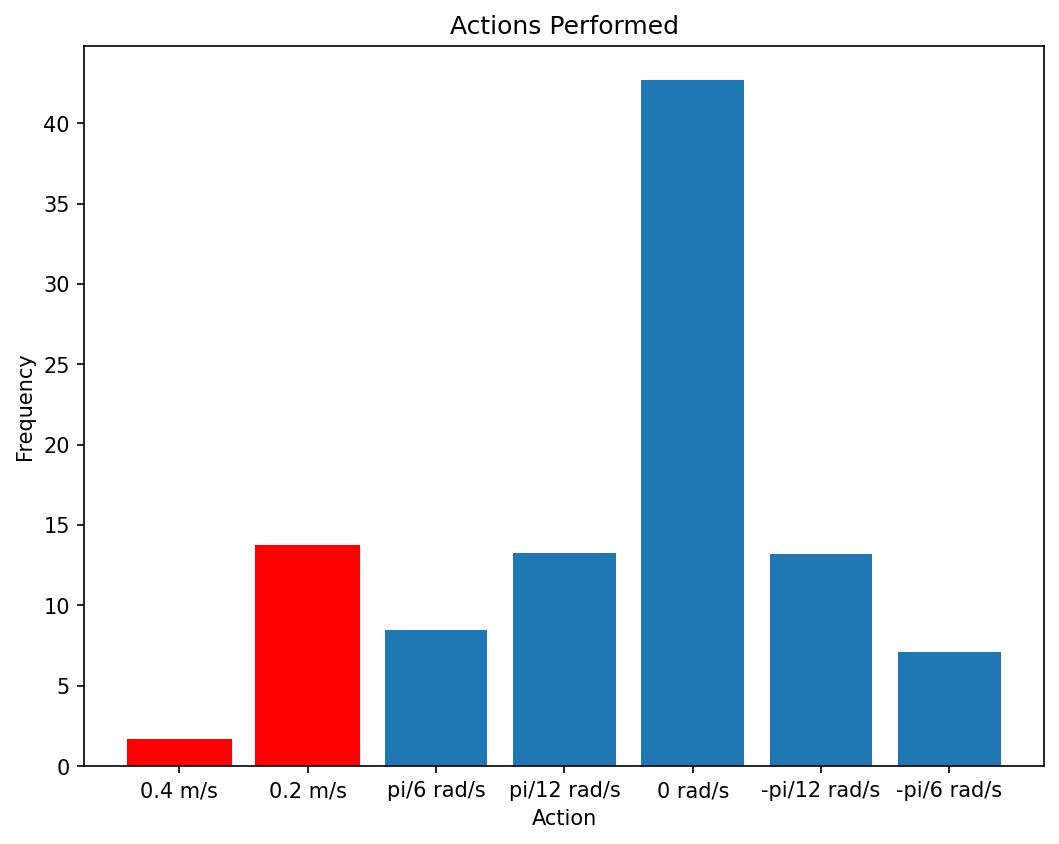}
        \caption{PositiveRainbow}
        \label{fig:navact2v2}
     \end{subfigure}
        \caption{Distributions of the chosen actions by the goal-oriented navigation agents during evaluation. NegativeRainbow shows more confidence by choosing the higher linear speed, even though both agents are rewarded by its choice.}
        \label{fig:navactions2}
\end{figure}

Although both agents were trained under the same restrictions for their choices of linear and angular velocities, there was a noticeable difference in the distribution of their chosen actions, evidenced in Fig. \ref{fig:navactions2}. NegativeRainbow preferred the highest linear speed and relied less on turning, displaying its better mastery of the task, while PositiveRainbow favoured a lower linear speed.

\begin{table}[h!]
\begin{center}
\caption{Evaluation performance of the goal-oriented navigation agents under the same circumstances. NegativeRainbow almost beat the environment by achieving a near-perfect goal-reaching rate.}
\begin{tabular}{c c c} 
\toprule
  Agent & Average Goal Reached Percentage & Average Steps \\ 
  \midrule
  NegativeRainbow & \textbf{96.9\%} & \textbf{79}  \\ 
  PositiveRainbow & 67.9\% & 173\\ 
  \bottomrule
\end{tabular}
\label{navevaltab2}
\end{center}
\end{table}

\begin{figure}
     \centering
     \begin{subfigure}[b]{0.49\textwidth}
         \centering
        \includegraphics[width=\textwidth]{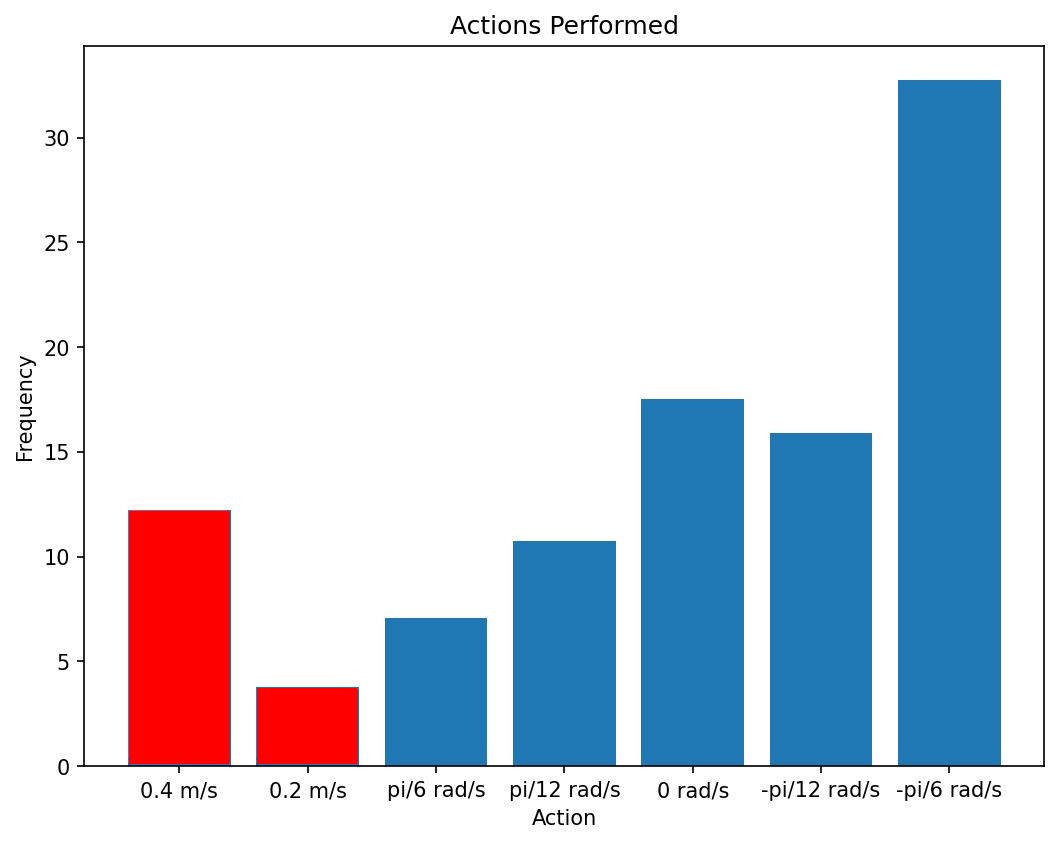}
        \caption{NegativeRainbow}
        \label{fig:navact1}
     \end{subfigure}
     \hfill
     \begin{subfigure}[b]{0.49\textwidth}
         \centering
        \includegraphics[width=\textwidth]{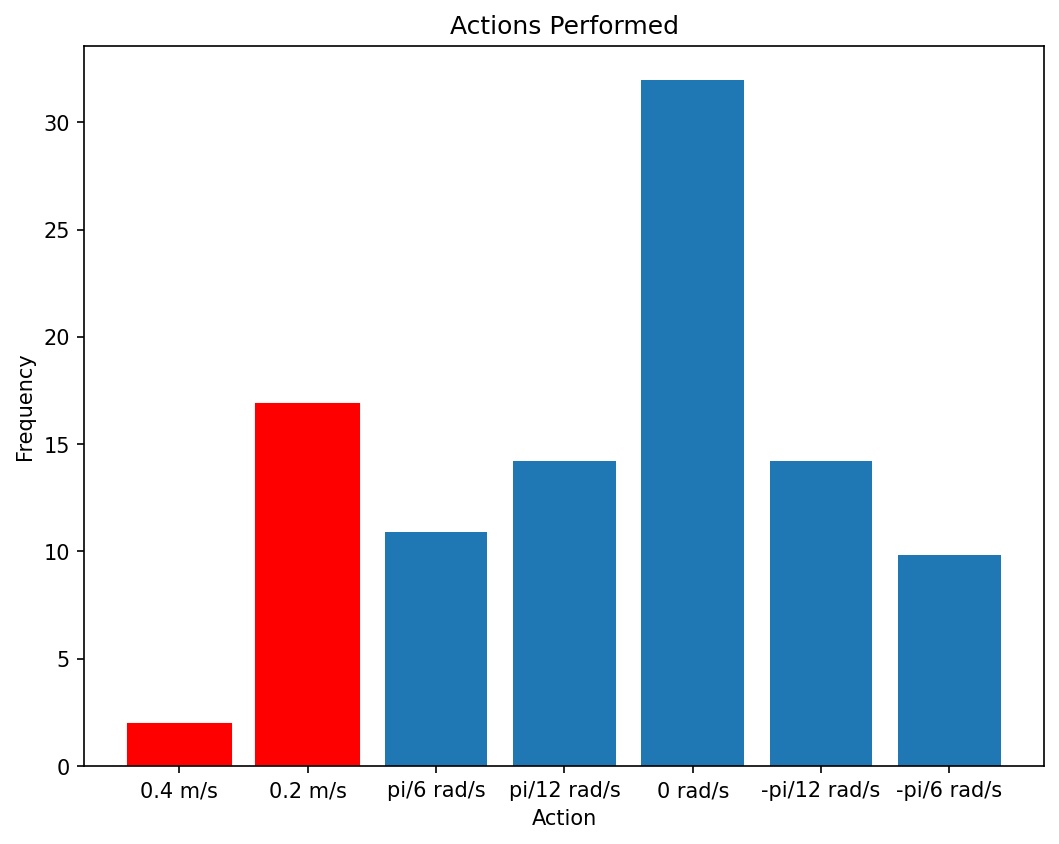}
        \caption{PositiveRainbow}
        \label{fig:navact2}
     \end{subfigure}
        \caption{Distributions of the chosen actions by the goal-oriented navigation agents during evaluation under different initial conditions. NegativeRainbow preferred to turn around, while PositiveRainbow kept its original distribution.}
        \label{fig:navactions}
\end{figure}

However, once the initial conditions are changed, there is a sharp decline in both agents' performance, as seen in Table \ref{navevaltab}, with them reaching the goal less than half the amount of times compared with the previous evaluation. It is also noteworthy that it almost took the agents the maximum number of steps to reach the goal. Their uncertainty was also reflected in their actions, shown in Fig. \ref{fig:navactions}, with both agents performing higher turning.

\begin{table}
\begin{center}
\begin{tabular}{c c c} 
\toprule
  Agent & Average Goal Reached Percentage & Average Steps \\ 
  \midrule
  NegativeRainbow & \textbf{35.5\%} & \textbf{200}  \\ 
  PositiveRainbow & 22.6\% & 228\\ 
  \bottomrule
\end{tabular}
\caption{Evaluation performance of the goal-oriented navigation agents under different initial conditions. The agent's average number of steps required to reach the goal almost reached the limit.}
\label{navevaltab}
\end{center}
\end{table}

 The lower performance led to the belief that either the state representation for the navigation task needed to be better or the NN ignored the polar coordinates. Furthermore, the agent seemed to learn to reach the goal by visually recognising the path in its training environment.

\begin{figure}
     \centering
     \begin{subfigure}[b]{0.49\textwidth}
         \centering
        \includegraphics[width=\textwidth]{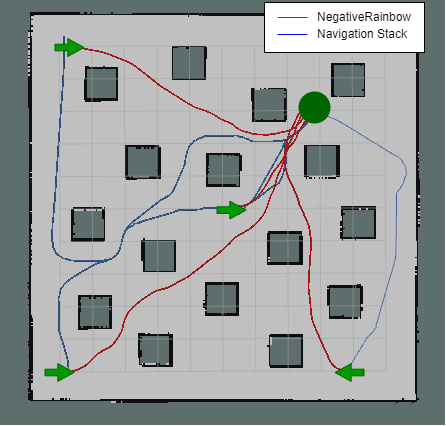}
        \caption{North-East Goal Position}
        \label{fig:mapa1}
     \end{subfigure}
     \hfill
     \begin{subfigure}[b]{0.49\textwidth}
         \centering
        \includegraphics[width=\textwidth]{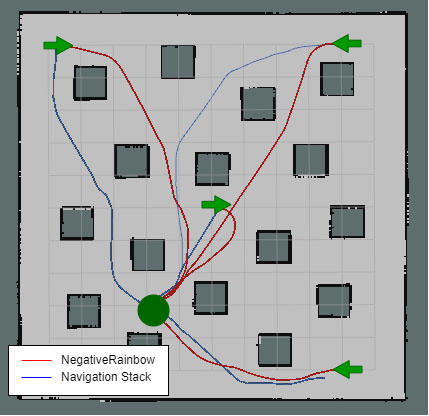}
        \caption{South-West Goal Position}
        \label{fig:mapa2}
     \end{subfigure}
        \caption{Comparison between NegativeRainbow's trajectories and the standard Turtlebot2 navigation stack, which uses Dijkstra's algorithm and DWA. Arrows indicate the starting configurations, and the circle is the goal position.}
        \label{fig:stackcomp}
\end{figure}

 Nonetheless, some NegativeRainbow's trajectories in its training environment, where it fulfilled its task almost entirely, were compared with the path computed by the standard navigation stack developed for the Turtlebot2 in Fig. \ref{fig:stackcomp}. The navigation stack uses Dijkstra's algorithm for the global path planner and DWA for local path planning. To compute the paths, it was required first to manually generate the cost map using the GMapping package. The RL agent's trajectories only required loading the trained policy but were registered on the same map for clarity.

Even if the agent's capability to generalize its knowledge was lacking, it seemed to approach the optimal behaviour in its training environment, as its trajectories were straight and shorter than using a path planner. NegativeRainbow preferred to navigate between the obstacles to reach the goal faster rather than planning a path around them.

\section{Discussion}

The best obstacle avoidance agent, SimpleRainbowL, reached below the twenty per cent collision rate during training, as seen in Fig. \ref{fig:obstrainv1}. It also achieved 41.83\% under different conditions, as evidenced in Table \ref{obsperf}, and 48\% in a different environment, reported in Table \ref{mazeperf}. It hinted that a simple reward function might be enough to fulfil a task if the state representation is adequate, in this case, the depth images.

Training with a larger image size yielded slightly better results but required more time to train for the same amount of episodes. And when the agents were evaluated in a different environment, the size of the images they used appeared to influence their results more, as seen in Table \ref{mazeperf}. More experiments would be required to validate these assumptions or verify if the standard image size was too small.

The agents with the behaviour reward function took longer to lower their collision rates, as seen in Fig. \ref{fig:obstrain}, because they had to consider the constraints imposed by their reward function when optimising their policy. Nonetheless, rewarding the value of the linear velocity and penalising the angular velocity achieved the expected result, as the agents preferred to move faster and turn less. This behaviour may not be ideal for the obstacle avoidance task, where the agent must prioritise avoiding collisions rather than moving fast. Still, it served as a proof of concept and a basis when designing the reward function of the goal-oriented navigation task where reaching the objective faster was preferred.

When switching to the goal-oriented navigation task, the distance and angle to the goal were added to the state representation to reward the agent for moving closer to the objective. Nonetheless, the agent also had to use depth images to avoid obstacles. The best agent, NegativeRainbow, achieved a 96.9\% rate of reaching the goal in its training environment and 35.5\% under different conditions, seen in Tables \ref{navevaltab2} and \ref{navevaltab} respectively. Its success during training seemed to be due to the use of depth images as part of the state representation, as the drop in performance during evaluation indicated that the agent could not constantly localise its goal. Furthermore, the analysis of the agent's trajectories during motion seemed to imply that it almost reached the optimal behaviour in its training environment. However, further study and measurements would be required to confirm it.

It was noteworthy that NegativeRainbow almost always reached the goal in its training environment while also avoiding obstacles, suggesting that using depth images might be enough to learn the task in a specific environment. However, the lack of solid generalisation to different conditions makes it impractical for use in different environments. The results imply that switching from a simpler task to a more complex one requires more consideration than adding the additional information required. 

In the goal-oriented navigation task, the penalty at each time step improved the agent's performance, seen in Table \ref{navevaltab2}, as the agent was urged to reach the destination as fast as possible. On the contrary, a reward at each time step might have instigated the agent to accumulate reward by navigating rather than by reaching the goal, as the agent still appeared to increase its average return during training, which can be noticed in Fig. \ref{fig:navtrain}.

In all cases, the Rainbow DQN algorithm achieved better results than the D3QN algorithm. The improved exploration mechanism provided by the noisy network, the consideration of additional time steps in the computation of the n-step return, and the better estimations provided by the use of the distribution of the rewards aided the agents in accomplishing both tasks. Similar to the original paper, where Hesselt et al. \cite{rainbowdqn} demonstrated a significant improvement in the results of Rainbow over previous variations of the DQN algorithm, the agents trained with the Rainbow DQN algorithm reached significantly more goals and collided less with obstacles than those trained with D3QN. This supports the idea of using variations of DRL algorithms with additional improvements to achieve better results rather than clinging to the most popular ones.

Finally, all agents trained seemed to learn their tasks with varying degrees of success, as their average reward kept increasing and the loss of their algorithm decreased during their training. Longer training sessions could increase the agents' performance even further.

\section{Conclusions}
This research project involved implementing a DRL approach for different robot navigation-related tasks. The proposed methods achieved a 41.83\% collision rate for the obstacle avoidance task and a 96.9\% target-reaching rate for the goal-oriented navigation task in their training environments. However, their lower performance during evaluation suggests that further work is required to achieve optimal behaviour.

The experimental work suggests that the improved exploration, more informed updates and better estimations of the Rainbow DQN allowed it to reach more targets and collide less during training than the D3QN agents. The results support the idea that, much like its comparison with the previous variations of the DQN method in their original domain of Atari games, Rainbow DQN might also perform better at navigation-related tasks. This could lead to improvements in existing works or as an idea to consider when designing a new DRL approach in the same field.

To perform the goal-oriented navigation task, the agent was provided additional information to measure how close it was to the goal compared to the design of the obstacle avoidance task. The trained agent seemed to succeed at the task in its training environment with a 96.9\% goal-reaching rate but only achieved 35.5\% under different conditions, seemingly learning the specific path to the goals during training. The results suggest that the transition from the obstacle avoidance task to the goal-oriented navigation task could not be accomplished with the parameters added for the agent's localisation and that further study should be performed about the state representation or balance of the weight of each data source.

Finally, a behaviour was induced in an obstacle avoidance agent by placing penalties based on its linear and angular velocities in the reward function, which led to the robot preferring to move faster and avoid turning. Still, it avoided fewer obstacles than using a simpler reward function with the same amount of training, suggesting that it required more time to learn its task. In the case of the goal-oriented navigation task, a penalty at each time step encouraged an agent to reach the target faster and more consistently than using a reward function that could grant positive values. Similar constraints could also be implemented for other navigation-related tasks, but a trade-off between training time and performance might still apply.

\bibliographystyle{abbrv}
\bibliography{reference}

\end{document}